\documentclass[preprint,12pt]{elsarticle}

\usepackage{acro}
\usepackage{graphicx}
\usepackage{caption} 
\usepackage{subcaption} 
\usepackage{pdflscape}
\usepackage{tabu}
\usepackage{longtable}[=v4.13]
\usepackage{float}
\usepackage{graphicx}
\usepackage{subcaption}
\usepackage{siunitx}
\usepackage{xcolor}
\usepackage{lineno}
\usepackage{booktabs}
\usepackage[normalem]{ulem}
\usepackage{rotating}
\usepackage{lineno} 

\begin{document}

\begin{frontmatter}




\title{Enhancing Machine Learning Performance through Intelligent Data Quality Assessment: An Unsupervised Data-centric Framework}

\author[inst1]{Manal Rahal}
\affiliation[inst1]{organization={Department of Mathematics and Computer Science, Karlstad University},
            addressline={Universitetsgatan 2}, 
            city={Karlstad},
            postcode={65188}, 
            country={Sweden}}

\author[inst1,inst2]{Bestoun S. Ahmed}
\affiliation[inst2]{organization={Department of Computer Science, Faculty of Electrical Engineering, Czech Technical University},
            city={Prague},
            postcode={16627}, 
            country={Czech Republic}}

\author[inst3]{Gergely Szabados}
\affiliation[inst3]{organization={Department of Engineering and Chemical Sciences, Karlstad},
            addressline={Universitetsgatan 2}, 
            city={Karlstad},
            postcode={65188}, 
            country={Sweden}}

\author[inst3]{Torgny Fornstedt}

\author[inst3]{Jörgen Samuelsson}

\begin{abstract}
Poor data quality limits the advantageous power of Machine Learning (ML) and weakens high-performing ML software systems. Nowadays, data are more prone to the risk of poor quality due to their increasing volume and complexity. Therefore, tedious and time-consuming work goes into data preparation and improvement before moving further in the ML pipeline. To address this challenge, we propose an intelligent data-centric evaluation framework that can identify high-quality data and improve the performance of an ML system. The proposed framework combines the curation of quality measurements and unsupervised learning to distinguish high- and low-quality data. The framework is designed to integrate flexible and general-purpose methods so that it is deployed in various domains and applications. To validate the outcomes of the designed framework, we implemented it in a real-world use case from the field of analytical chemistry, where it is tested on three datasets of anti-sense oligonucleotides. A domain expert is consulted to identify the relevant quality measurements and evaluate the outcomes of the framework. The results show that the quality-centric data evaluation framework identifies the characteristics of high-quality data that guide the conduct of efficient laboratory experiments and consequently improve the performance of the ML system. 
\end{abstract}

\end{frontmatter}

\section{Introduction}\label{sec:intro}
Machine learning (ML) as a growing innovative field has proven successful in numerous domains \cite{Schmidt2019}. In recent years, ML programs have evolved into data-driven software systems with a complex yet defined pipeline of operations. Given that an ML software system is based on learning patterns from training data, the first operation in the pipeline typically includes data preparation \cite{Kunft2019}. Once the data are clean and of high quality, the training and evaluation operations of the ML model are carried out with confidence. Although research shows that data quality (DQ) has a direct impact on the outcomes of an ML software system, the evaluation of DQ has been identified as the greatest challenge for ML practitioners (\cite{Amershi2019,Chen2021}).

Poor DQ leads to weak learning from the training data and threatens the generalization power of the model. Therefore, optimizing the performance of the ML software systems could not be achieved without optimizing both the ML model and the DQ \cite{Sambasivan2021}. In the literature, DQ evaluation refers to the identification of data dimensions and metrics to measure the importance of data. However, the assessment of DQ is not straightforward, as it is highly dependent on the nature of the application \cite{Gudivada2017}. Therefore, DQ has been addressed from different perspectives in different applications and domains.

In the field of analytical chemistry, data-driven research, including the application of ML, has a long history \cite{Schmidt2019}. One of the most popular applications of predictive ML models in this field is the prediction of retention time ($t_\mathrm{R}$) in chromatography experiments. During a chromatography experiment, the constituting compounds of a complex mixture undergo separation. The time a compound spends in a chromatography column is called $t_\mathrm{R}$. Chromatography-generated raw data could include inconsistent recordings. The source of DQ inconsistency can be attributed to the performance of the chromatographic equipment, experimental conditions, and other external factors. Given that ML operates under the assumption of clean data, DQ should be addressed before it is fed into the ML model in the larger ML software system \cite{Atla2011}. In many cases, the evaluation of DQ requires manual intervention from chemical scientists, which is often difficult to implement and time-consuming. 

To address this gap, this paper proposes a quality-centric data evaluation framework for a fast and simple multi-stage approach to DQ evaluation. The proposed framework is built on unsupervised learning to address quality issues before training and evaluating the ML model. The advantage of unsupervised learning is that the model learns the characteristics of the new data and updates the rules accordingly. In other words, it is expected to be more flexible to data drifts and changes. Using a systematic and automated DQ evaluation framework in the early stages of the ML software system pipeline improves the quality of training and testing data, thus enhancing the prediction of $t_\mathrm{R}$. The framework presents scientists with a practical method to narrow the search space for poor DQ records. For example, it enables analytical chemists to identify the highly probable overlapping peaks so they can identify the individual compounds. As a result, analytical chemists gain insight into the characteristics of high-quality data, facilitating the design and implementation of more efficient experiments. In other words, reduces the experimental burden in terms of time and money.

This paper aims to present a quality-centric data evaluation framework for large datasets, that requires minimal human intervention. The proposed framework is validated on three datasets, consisting of anti-sense oligonucleotide (ASO) samples and their experimentally observed retention times. The design and deployment of the framework in a real-world use case provide three main contributions: (1) A general-purpose DQ evaluation framework is developed to improve the performance of ML software systems. (2) Unsupervised and predictive ML methods are integrated to deduce the characteristics of high-quality data. (3) The DQ evaluation framework is validated on a real-world use case. 

The remainder of the paper is organized as follows.  In Section \ref{sec:relatedwork}, relevant and different perspectives on DQ issues and frameworks are highlighted and discussed. In Section \ref{sec:DQframework}, the motivation to select quality measurements and the unsupervised learning method are discussed. Section \ref{sec:implementation} describes the implementation of the quality-centric framework in the case study of chromatography data and presents the findings. The findings are analyzed and discussed according to three research questions (RQs) in Section \ref{sec:discussion}. Finally, Section \ref{sec:conclusion} concludes the study and the results.

\section{Background and Related Work}\label{sec:relatedwork}
This section gives an overview of common data issues in ML software systems and discusses the work related to developing DQ frameworks in the literature. Given that the case study is applied to an application in the field of analytical chemistry, this section also presents common DQ issues and handling in the context of chromatographic data.

\subsection{Data Issues in ML Software Systems}
Data understanding and preparation are often the least interesting stages in the ML software pipeline. However, to ensure proper learning, DQ must be handled \cite{Corrales2018}. During this phase, the features are encoded, the data are checked for null values, outliers, and statistical correlation among the independent variables, and the distribution of the feature values is visualized. The importance of data pre-processing comes from the critical impact of the data on the performance of the ML software system. 

The literature studying this topic separates the data issues observed in information systems from big data issues. In \cite{Wand1996}, the authors identified the data problems observed in information systems, such as information loss, ambiguity, meaningless, or incorrect data. The categorization is based on the true representation of an information system in the real world and the fact that data issues arise due to representation deficiencies \cite{Wand1996}. With the evolution of big data and AI, researchers have begun to investigate DQ issues beyond classical database operations and have begun to look at big data challenges \cite{Gudivada2017}.

Despite data issues dating back to the early days of computing, DQ is still considered an interesting research topic \cite{Gudivada2017}. In particular, the challenges emerging from big data. Although big data has great potential for the advancement of technologies, at the same time, it presents many challenges that arise from its properties \cite{Katal2013}. Most research extracts data issues from the characteristics of the big data itself. For example, Fan \emph{et al.} identifies the challenges of big data in complex heterogeneity, high dimensionality, noise accumulation, spurious correlation, incidental endogeneity, and measurement errors \cite{Fan2014}. Another research considered heterogeneity and incompleteness, scale, and timeliness as data-related issues to be addressed \cite{Ammu2013}. In another study, the authors present a comprehensive mapping of data issues into three main categories \cite{Thabet2015}. The first category includes data issues related to the characteristics of big data, such as volume, variety, velocity, veracity, volatility, and variability. The second group is related to the challenges of data processing from collection to the application of ML. The third category pertains to data management issues, including data security, privacy, and ethical concepts.  

There exist different perspectives to investigate DQ in ML systems. Given that the data used in ML systems are often large, it automatically inherits the data problems mentioned in \cite{Thabet2015} and \cite{Fan2014}. Although many of the mentioned data issues are cross-cutting, the quality of the data depends to a great extent on the nature of the application being studied. Therefore, in our use case, the quality measurements are specific to the chromatography dataset and are deduced from experimental recordings. The deduced measurements fall mostly under the veracity and variability issues of big data.

\subsection{Systemic Approaches to DQ Evaluation in the Literature}\label{sec:DQapproaches}
The concept of DQ greatly depends on the nature of the application under study. Therefore, the definition and assessment of DQ is a complex concept \cite{Gudivada2017}. As such, there are different definitions of DQ in the literature. For example, DQ could refer to the measurement of incorrect or missing data \cite{Firmani2016}, or represent the suitability of a given dataset, including the features, for a specific use case (\cite{Poon2021}, \cite{nikiforova2020}). In this paper, we define the DQ as the characteristics of the data that fit the purpose of building high-performing ML systems.

DQ validation is an influential requirement for a reliable ML system \cite{Foidl2019}. Given its dependence on the use case, researchers from different perspectives have handled the concept of DQ. However, many of the DQ frameworks in the literature are based on database management concepts that aim to solve generic DQ problems \cite{taleb2021}. One of the first DQ frameworks proposed testing whether the data are complete, unambiguous, meaningful, and correct \cite{Wand1996}. While in \cite{WangStrong1996}, authors go beyond precision and present a hierarchical framework that categorizes the attributes of DQ into four groups: intrinsic, contextual, representational, and accessible. According to \cite{WangStrong1996}, high-quality data are defined as intrinsically good, contextually suitable for the task, clearly represented, and accessible. In \cite{Juneja2019}, the authors address the quality of raw data in the pre-processing stage and apply it to a conceptualized weather monitoring and forecasting application as a case study. The idea is to fix as many data issues as possible before training and evaluating ML models \cite{Juneja2019}. \cite{Foidl2019} presented a risk-based data validation approach in ML systems inspired by the famous risk-based approach in SE testing. In the risk-based approach, features are presented as risk items where the risk of poor DQ for each feature is estimated. Two important factors are calculated under this approach, the probability that a feature is of low quality and the impact of this feature on the performance of the ML system. This approach presented a new perspective to DQ evaluation, but it was not validated in a real-world application. 

In Literature, some of the DQ frameworks are, by design, limited to specific applications or specific types of data. For example, \cite{Byabazaire2022} proposed a real-time DQ assessment to integrate trust metrics into the Internet of Things (IoT) data cycle. The framework was tested using data from real-time IoT sensors \cite{Byabazaire2022}. The results and analysis showed that the trust metric could be a good DQ metric in the context of IoT data \cite{Byabazaire2022}. In another study, an exhaustive review of DQ assessment and improvement methods is presented; however, it is limited to specific applications \cite{batini2009}. In \cite{Bayram2023}, a DQ scoring framework is presented for production data. The framework yields an aggregated score based on five quality dimensions: accuracy, completeness, consistency, timeliness, and skewness. In \cite{Chen2021}, the authors propose quality attributes that are most important for deep learning. The three quality attributes finely selected by Chen \emph{et al.} are comprehensiveness, correctness, and variety, which they redefined to fit deep learning applications. As part of the study, they conducted experiments to investigate how noisy data could lead to a false improvement in ML performance \cite{Chen2021}. The results of the experiments performed showed a strong correlation between DQ and ML performance \cite{Chen2021}. Similar preliminary research on deep learning has been conducted in \cite{Ding2019} and \cite{Chen2019}. 

A subset of the approaches were mainly guided by the type of data and not the task. For example, the quality of linked data was investigated in multiple research-based frameworks, such as in the Luzzu Quality Assessment Framework \cite{Luzzu2016}. The framework provides a 22-dimensional library of quality metrics that can be used to assess the fitness of the dataset for use for a specific user-defined task. Another study focusing on linked data proposed a five-step assessment framework that detects the root causes of DQ violations and provides improvement recommendations \cite{Nayak2021}. However, these frameworks were shown to be limited to the assessment of metadata. On the other hand, the quality of open data also attracted the attention of researchers, such as in \cite{Krasikov2023} where the authors proposed methodological guidance for screening, assessing, and preparing open data for business scenarios in an enterprise setting. The authors argue that “fitness for use” could apply to open data \cite{Krasikov2023}.

In the context of big data, a large DQ management framework that aims to address end-to-end DQ throughout the entire life cycle of big data is proposed in \cite{taleb2021}. The framework captures quality requirements, attributes, dimensions, scores, and rules and quantifies the scores for quality dimensions. However, the authors did not test the framework using real-world data. 

As observed, understanding the quality of the data is realized through many data-based or application-based approaches, with the most commonly measured dimensions being completeness, timeliness, and accuracy \cite{Cichy2019}. The success of any approach is correlated with finding interesting characteristics of the data that can be transformed into quality measures \cite{Patel2023}. One of the interesting methods to achieve this is through the application of unsupervised ML. In this paper, we propose a general framework that relies on the use of unsupervised learning to evaluate DQ using pre-defined quality measurements. As a result, the high-quality characteristics that make a high-performing ML system are deduced.

\subsection{DQ in Ion-pair Liquid Chromatography}

\begin{figure}
  \centering
    \includegraphics[scale=.2]{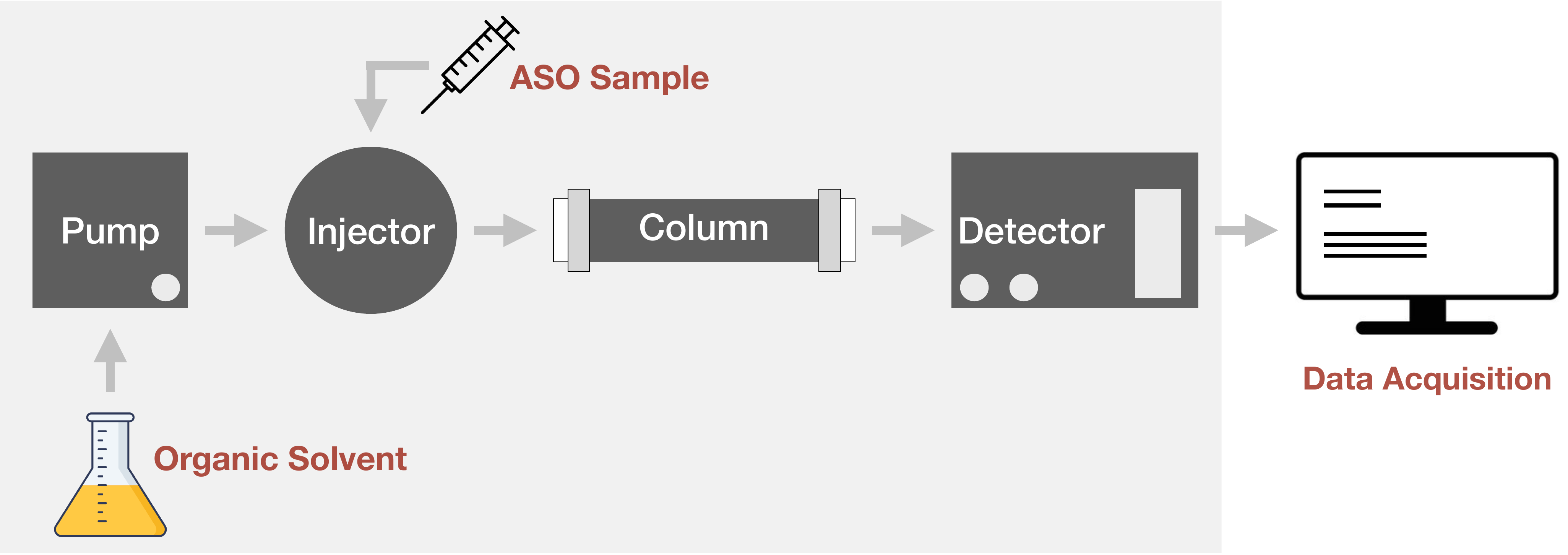}
    \caption{Separation process using liquid chromatography and a spectrometry detector. }\label{fig:IPLC}
\end{figure}

Chromatography, in general, is an efficient method used in the separation of a range of samples, including drugs, food, air and water samples, and much more. Ion-pair Liquid Chromatography (IPLC) is a chromatography technique that allows one to separate complex molecular mixtures in particular charged molecules \cite{ENMARK2022}. The method includes injecting the sample, including a mixture of compounds, into the column where the separation occurs, as shown in Figure \ref{fig:IPLC}. As the separated compounds exit the column, they are detected by the detector as signals. The information collected on the detected signals provides insight into the success of the separation. In our case study, a mixture is a group of ASO compounds separated from the impurities by the IPLC method. An ASO compound is a combination of the nucleotides adenine (A), thymine (T), cytosine (C), and guanine (G). An ASO sequence could be modified by other atoms, such as sulfur, known as phosphorothioation. The importance of ASO sequences comes from their interesting potential in the treatment of diseases that are not targeted by classical medicine \cite{Thakur2022, FORNSTEDT2023}. 

While the compounds are eluting from the IPLC system, the data describing the detected signals are recorded. A sample of the collected data is shown in Table \ref{tab:signalsdata}, where each row refers to an ASO compound and its peak characteristics. The characteristics of a peak include the signal-to-noise ratio (SNR), $\Delta$t$_R$, which calculates the difference in $t_\mathrm{R}$ between two experiments, and the area under the peak. The resulting peaks are typically visualized in a chromatogram, which is later inspected and analyzed by analytical chemists.

A chromatogram is a graphical representation of the detected signals over the separation time. Analytical chemists spend a substantial amount of time controlling the quality of the data visualized data in a chromatogram \cite{Felinger1998}. Data pre-processing in this context includes removing the undesired experiments from the obtained signals \cite{YANG2019}. A typical pre-processing procedure includes noise removal and baseline correction \cite{YANG2019}. Furthermore, inspecting the properties of the peaks, focusing mainly on the t$_\mathrm{R}$, shape, and resolution peaks \cite{Felinger1998}. In particular, inspecting the cases of overlapped peaks, shifted $t_\mathrm{R}$, and peaks with low SNR \cite{JOHNSEN2017}. Chromatogram inspection and analysis are crucial since accurate peaks lead to sound quantitative and qualitative analysis. 

\begin{table}
\setlength{\tabcolsep}{5 pt}
\centering
\caption{Data collected from detected signals in IPLC.}
\label{tab:signalsdata}
\begin{tabular}{lllllll}
\toprule
\textbf{Sequence}& \textbf{$\Delta$t$_R$} & \textbf{SNR} & \textbf{Skewness} & \textbf{Peak area}\\
\midrule
AAAAAAAAAAAAAAAAAAAA & -0.01 & 792.68 & 1.21 & 22443.81\\
CACGTGACTATG& $-0.00$ & 1893.19&	1.16& 52434.17 \\
A*C*G*T*G*ACTATG-P=O & $-0.03$&	18.36&	1.03&	239.42\\
A*TTAGAA*T*T*A & $-0.01$ &	61.09 &	1.18 & 699.10\\
\bottomrule
\end{tabular}
\end{table}

Chromatogram analysis is a critical step in correcting the data problems faced during the data acquisition phase. A chemist generally looks for significant peaks and observes failed experiments. The analysis of the data is mostly conducted manually with the assistance of statistical and mathematical tools. Still, it is often complex and time-consuming, as all resulting peaks must be checked. Therefore, the interest in using ML methods to facilitate data analysis is increasing.

In the literature, ML has been used for various peak quality control applications such as peak selection, integration, and annotation \cite{Liebal2020}. Regarding peak picking, multiple ML models were developed to process and select chromatogram peaks, including deep learning approaches \cite{Liebal2020}. In particular, using convolutional neural networks as in \cite{RISUM2019,ZHANG2019}. However, in this paper, we do not focus on a specific aspect of DQ but instead use unsupervised learning to learn about all potential data issues in the space of the selected quality measurements. 


\section{The proposed Quality-Centric Data Evaluation Framework}\label{sec:DQframework}
\begin{figure*}
  \centering
    \includegraphics[scale=.22]{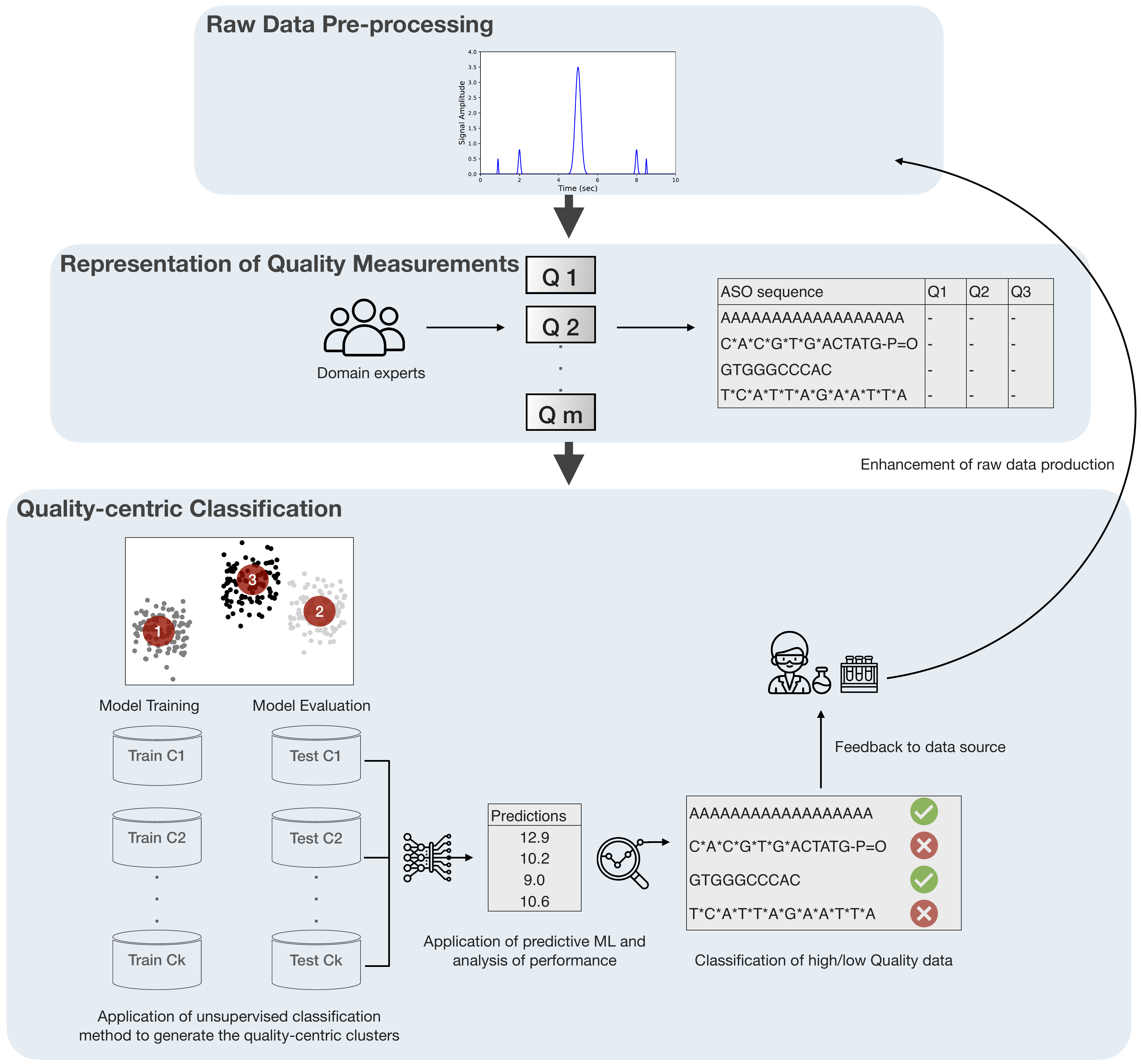}
    \caption{Proposed quality-centric data evaluation framework. The pipeline starts with raw data pre-processing, followed by quality measurements defined by domain experts ($Q1-Qm$). The framework then applies unsupervised classification to generate quality-based clusters, which undergo ML model training and evaluation. Results feed back to data source controllers to enhance future data collection and quality.}\label{fig:framework}
\end{figure*}

This section presents the methodology for implementing the quality-centric data evaluation framework. The output of the framework is a classification of the input data according to the designed quality measurements. The framework is built on general-purpose ML methods, which allows it to be applied in different domains. The resulting understanding of the quality of the input data can be used in various tasks depending on the nature of the business. For example, from a modeling perspective, high-quality data is fed to the ML software system, and lower-quality data could be rechecked by scientists for further analysis. In addition, low-quality data could be further analyzed to understand why certain chemical modifications can degrade or boost the performance of an ML software system. In this paper, the implementation of this framework reduces the time it takes scientists to check poor input data, as it reduces search space.
The overall phases of the framework are shown in Figure \ref{fig:framework}. Once the raw data are received, the domain experts, along with the ML practitioners, decide on the quality measurements that best fit the nature of the application. The quality measurements are then represented in a dataset where they serve as independent variables for the later application of the unsupervised learning method. The resulting dataset is then sufficiently pre-processed and fed to an unsupervised ML model to generate \emph{k} quality-sensitive clusters, where \emph{k} represents the number of clusters. \emph{k} can be identified through multiple methods such as the elbow method. After the clusters are generated and plotted, a hyper-tuned ML model, using the grid search method, is applied to each of the \emph{k} clusters' data, where the target variable is predicted. This stage is critical, as it allows one to understand the performance of the ML model on the different clusters. Per cluster, ML performance metrics on the train and test sets are recorded and analyzed. During the analysis stage, insights on the level of quality of the data are learned and fed back to data source controllers. The continuous feedback results in more efficient business operations that lead to improved ML systems.

\section{Implementation and Evaluation}\label{sec:implementation}

In this section, we adopt the quality-centric data evaluation framework for the case study of predicting $t_\mathrm{R}$ of ASO compounds including defining the relevant DQ measurements. The proposed framework is applied to three ASO datasets collected through IPLC experiments, described in Section \ref{sec:statistics}. Based on the experimental setup, we focus on answering three RQs:
\begin{itemize}
\item (RQ\textsubscript{1}) How can unsupervised learning be effective in classifying data records into different quality levels?
\item (RQ\textsubscript{2}) How do we transform the results of the application of an unsupervised DQ evaluation framework into explainable insights that improve the performance of the ML software system?
\item (RQ\textsubscript{3}) How can we validate the results of a DQ evaluation framework built on unsupervised learning?
\end{itemize}

\subsection{Defining DQ Measurements}\label{measurements}
Quality can be defined as a set of multiple dimensions in a specific context, where each dimension represents an aspect of the DQ and can be inherited from one or more measurements \cite{Poon2021}. The task of defining quality measurements is critical to the successful implementation of the framework. In principle, it is considered good practice to re-engineer the raw data before incorporating them into an ML software system \cite{Feinberg2017}. Otherwise, important patterns and information may be missed. At the same time, data representation is strongly associated with the goal of unsupervised learning \cite{Jain2010}. In clustering, even without natural clusters, data objects could be assigned to the same group \cite{Jain2010}. Therefore, the curation and design of quality measurements is a critical task. With the support of domain experts and analytical chemistry theory, relevant DQ measurements are selected, that describe the characteristics of the signal detected by the detector. As a result of rigorous data analysis, four crucial quality measurements were defined. Since data quantification is needed for data analysis, the derived measurements are quantified before application. The derived measurements represent the quality of the ASO data in the three datasets, which serve as input to the unsupervised ML method. In addition to the curated quality measurements, we also added the length of the ASO sequence and the amount of sulfur modification. In the following sections, the derived quality measurements are described in detail.

\subsubsection{SNR}
\begin{figure*}
  \centering
    \includegraphics[scale=.39]{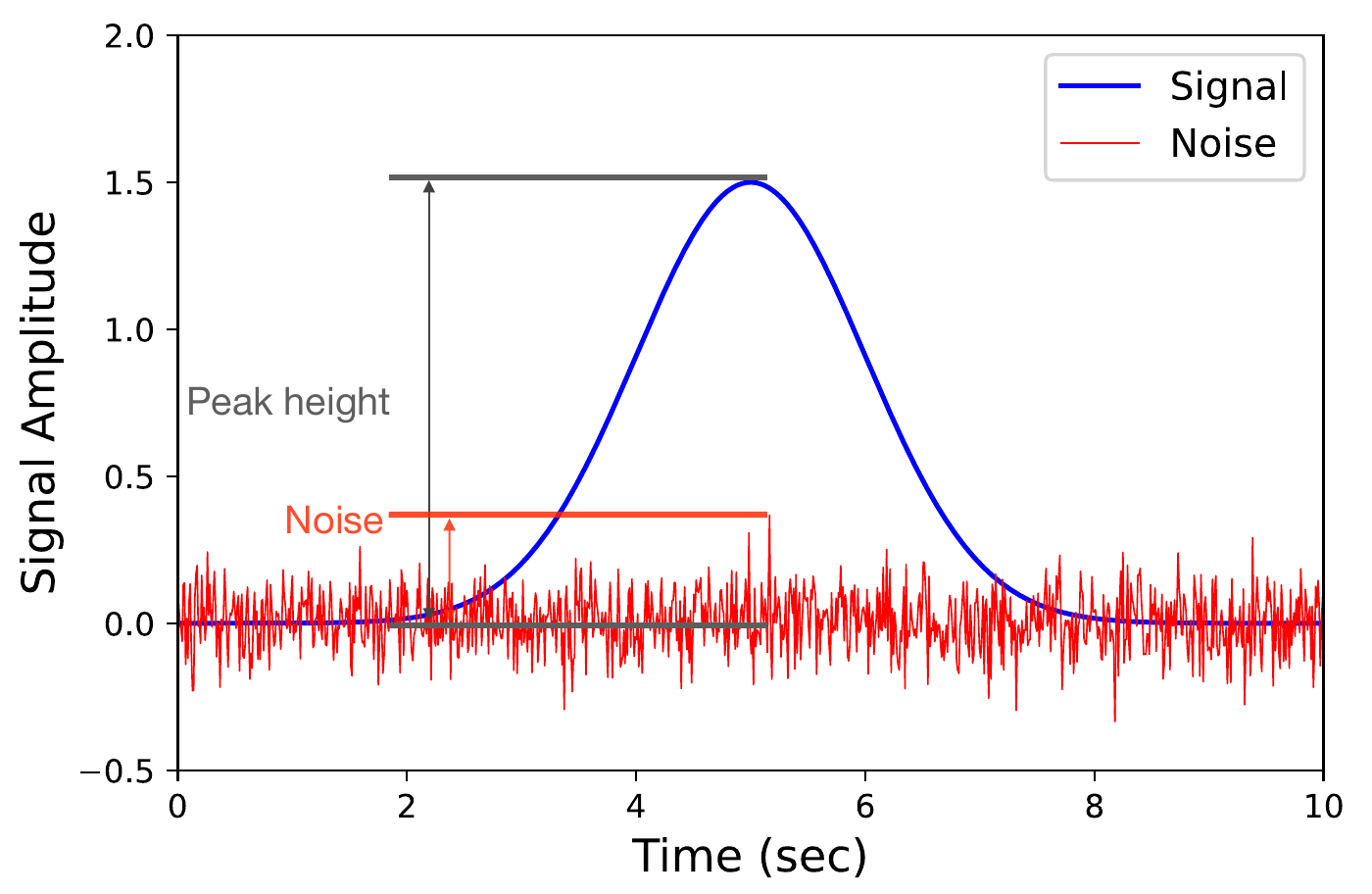}
    \caption{Chromatographic signal where the noise and the peak height are labeled.}\label{fig:snrsignal}
\end{figure*}

A chromatographic signal is obtained after an ASO sample passes through the column and is detected by the detector. The magnitude of the intensity of the signal at a specific time is a combination of the baseline signal of the chromatogram $B(t)$, the peak signal $P(t)$, and the noise signal $N(t)$, where noise commonly refers to unwanted fluctuations in the experimental system \cite{Jia2022}. An SNR is generally defined as the amount of noise relative to the main signal, as illustrated in Figure \ref{fig:snrsignal}. Hence, the chromatographic signal $Y(t)$ can be expressed as:
\begin{equation}
Y(t) = B(t) + P(t) + N(t)
\label{eq:signal}
\end{equation}



In chromatography, measurements depend on the baseline identified when the chromatogram is idle \cite{RUPPRECHT2022}. The higher the SNR, the better the quality of the signal. Assuming its definition, SNR is the first quality measurement to be considered in this paper.

\subsubsection{Delta $t_\mathrm{R}$ ({$\Delta$t$_R$})} For a given sample where two replicates were conducted, the $t_\mathrm{R,1}$ is obtained from the first chromatography experiment run. The $t_\mathrm{R,2}$ is obtained from the second run of the experiment. Assuming that all chromatographic conditions are the same, $\Delta$t$_R$ measures the fluctuation in t$_\mathrm{R}$ among repeated experiments. $\Delta$t$_R$ is calculated for each sample by subtracting $t_\mathrm{R,2}$ from $t_\mathrm{R,1}$. $\Delta$t$_R$ is the second selected measurement.

\subsubsection{Peak skewness} The symmetry of a signal peak in chromatography is an extremely important factor \cite{Papai2002}. In reality, the experimentally obtained peaks are asymmetric and commonly tailing \cite{Patrik2015}. In the literature, there are many methods to determine the skewness \cite{Papai2002}. A typical representation of skewness is the ratio $ W_{R}$ $/$ $ W_{L}$ at a selected peak height \cite{POOLE1991}. In the context of chromatography data, skewness, whether in the form of tailing or fronting peaks, is usually undesirable. Rather, the more symmetrical a peak is, the higher the quality of the data. In this case study, we calculate the skewness of a peak as in Eq.\ref{equation2}:


\begin{equation} Q_{s,0}(x) = \frac{W_{R}(x)}{W_{L}(x)} \quad \text{where} \quad x \in (0,1) \label{equation2} \end{equation}

\begin{figure*}
  \centering
    \includegraphics[scale=.6]{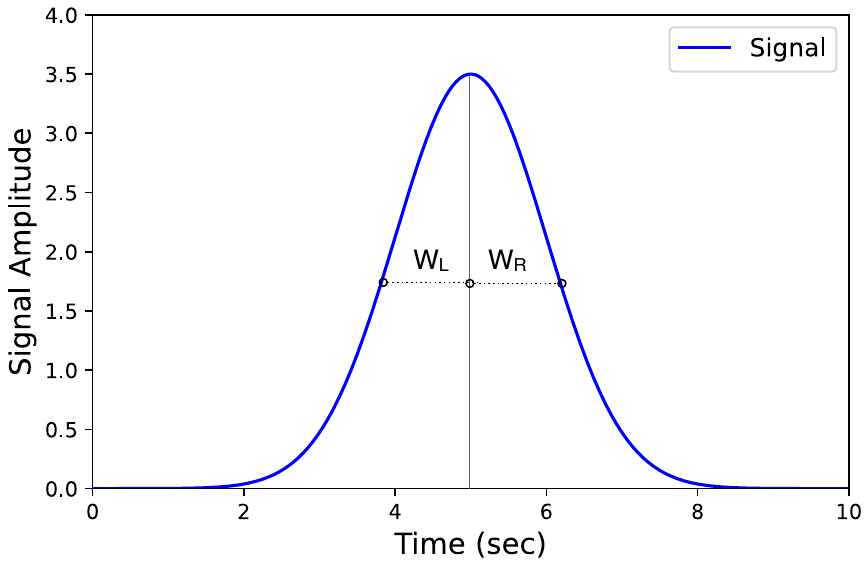}
    \caption{Symmetric peak, skewness = 1. Skewness is measured at x=0.5.}\label{fig:skewsignal}
\end{figure*}

$ W_{R} (x)$ and $ W_{L} (x)$ represent the horizontal distance to the line through the peak, at x=0.5 of the peak height as shown in Figure \ref{fig:skewsignal}. At $q_{s,0}$ =1, the peak is symmetrical. If $q_{s,0}$ $\in ]0, 1[ $, the peak is a fronting peak skewed to the left. Otherwise its is a tailing peak and $q_{s,0}$ $\in ]1, \infty[ $.

\subsubsection{Peak area} Another important factor in the chromatographic signal data is the peak area. In other words, it refers to the area under the curve. The area of the peak for a certain eluting compound is proportional to the amount of the compound reaching the detector.

\subsection{Description of ASO Datasets}\label{sec:statistics}
Three datasets were obtained by performing IPLC experiments on ASO samples to separate the full-length ASO compound from its impurities. Data were collected under three different chemical conditions, which resulted in three different datasets as shown in Table \ref{tab:datasets}. The change in the composition of the mobile phase, typically the organic solvent, is known as the gradient. The gradient allows for different levels of variability and noise in each dataset. The appropriate amount of mobile phase is critical to achieve an acceptable separation in a given experiment \cite{Patrik2015}. The three datasets have the same number of features but differ in the distribution of the independent variables and the target variable. Independent variables include the defined quality measurements, in addition to the ASO length and the amount of sulfur. In this case, the target variable is t$_\mathrm{R}$. The train set and the test set were collected from the same set of experiments, with all conditions unchanged, so we assume that both sets exhibit similar DQ characteristics. Since the data are collected from chromatography experiments, errors could occur during the experiment run-time, either during data collection or data storage.

\begin{table}
\setlength{\tabcolsep}{5 pt}
\centering
\caption{Total number of non-null instances in ASO datasets.}
\label{tab:datasets}
\begin{tabular}{llll}
\toprule
Dataset name & Dataset description & \#instances &\#features \\ 
\midrule
G1           & Gradient = 11 min   & 865         & 8\\
G2           & Gradient = 22 min   & 863         & 8\\
G3           & Gradient = 44 min   & 861         & 8 \\
\bottomrule
\end{tabular}
\end{table}

\begin{figure*}
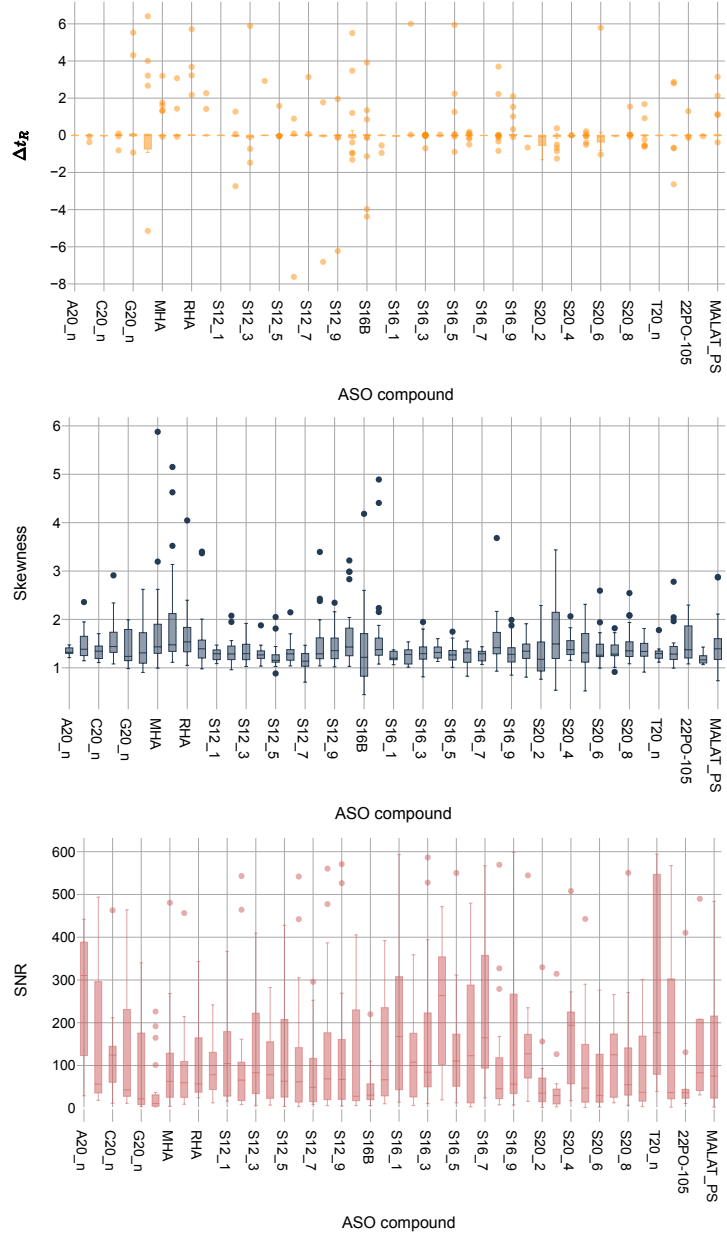

\centering
  \begin{subfigure}[b]{0.7\textwidth}
         \centering
         \includegraphics[width=\textwidth]{G11_dtR_product.pdf}
     \end{subfigure}
     \begin{subfigure}[b]{0.7\textwidth}
         \centering
         \includegraphics[width=\textwidth]{G11_skew_product.pdf}
     \end{subfigure}
     \begin{subfigure}[b]{0.7\textwidth}
         \centering
         \includegraphics[width=\textwidth]{G11_SNR_product.pdf}
     \end{subfigure}

    \caption{Distribution of quality measurements per ASO compound in G1 dataset: (a) {$\Delta t_R$} showing retention time variation, (b) skewness indicating peak symmetry, and (c) SNR demonstrating signal strength. The range of values has been adjusted\protect\footnotemark~for visualization purposes.}
    \label{fig:G1boxplots}
\end{figure*}

By applying the data quality measurements defined in Section \ref{measurements} to the three datasets in our use case, we analyze the variation in data distribution across these datasets. The results are presented using boxplots, showcasing the independent variables SNR, skewness, and $\Delta$t$_R$. As shown in Figures \ref{fig:G1boxplots}, \ref{fig:G2boxplots}, and \ref{fig:G3boxplots}, the highest variation in $\Delta$t$_R$ between the two experiments is found in the G3 dataset, while the lowest is found in the G1 dataset. The same applies to the SNR feature. Regarding the skewness feature, the mean in G3 is at least 26.22 times higher than in the other datasets.

\footnotetext{SNR boxplot is adjusted to include SNR values less than 600.}

\begin{figure*}
\centering
  \begin{subfigure}[b]{0.7\textwidth}
         \centering
         \includegraphics[width=\textwidth]{G22_dtR_product.pdf}
     \end{subfigure}
     \begin{subfigure}[b]{0.7\textwidth}
         \centering
         \includegraphics[width=\textwidth]{G22_skew_product.pdf}
     \end{subfigure}
     \begin{subfigure}[b]{0.7\textwidth}
         \centering
         \includegraphics[width=\textwidth]{G22_SNR_product.pdf}
     \end{subfigure}


 \caption{Distribution of quality measurements per ASO compound in G2 dataset: (a) {$\Delta t_R$} showing retention time variation, (b) skewness indicating peak symmetry, and (c) SNR demonstrating signal strength. The range of values has been adjusted\protect\footnotemark~for visualization purposes.}
    
    \label{fig:G2boxplots}
\end{figure*}
\footnotetext{SNR boxplot is adjusted to include SNR values less than 600.}

\begin{figure*}
\centering
  \begin{subfigure}[b]{0.7\textwidth}
         \centering
         \includegraphics[width=\textwidth]{G44_dtR_product.pdf}
     \end{subfigure}
     \begin{subfigure}[b]{0.7\textwidth}
         \centering
         \includegraphics[width=\textwidth]{G44_skew_product.pdf}
     \end{subfigure}
     \begin{subfigure}[b]{0.7\textwidth}
         \centering
         \includegraphics[width=\textwidth]{G44_SNR_product.pdf}
     \end{subfigure}

 \caption{Distribution of quality measurements per ASO compound in G3 dataset: (a) {$\Delta t_R$} showing retention time variation, (b) skewness indicating peak symmetry, and (c) SNR demonstrating signal strength. The range of values has been adjusted\protect\footnotemark~for visualization purposes.}
    
    \label{fig:G3boxplots}
\end{figure*}
\footnotetext{SNR boxplot is adjusted to include SNR values less than 600.}


\subsection{Motivation and Implementation of Selected Methods}\label{sec:methods}
Learning in the context of ML refers to finding patterns in the data \cite{Stefano2002}. The grouping or clustering of objects based on similarity is considered a fundamental exploratory method of learning \cite{Jain2010}. Clustering has been used mainly to detect anomalies and identify important features, perform natural classification of organisms, and compress the data into cluster prototypes. In the context of data science, clustering is simply the meaningful grouping of data objects \cite{IKOTUN2022}. Unsupervised clustering is a form of clustering that is applied to unlabeled data \cite{IKOTUN2022}. The clustering method is powerful by itself, but the results of the clustering task are often combined with a subsequent prediction task for higher explainability, as in our case study.

There are various clustering algorithms to analyze big data. Such as density-based clustering used in real-time data analysis and hierarchical clustering and incremental clustering for other applications \cite{Kolajo2019}. This study implements \emph{k}-means clustering, a hierarchical method, to group the ASO compounds into clusters according to pre-defined quality measurements. Two main advantages contribute to the selection of \emph{k}-means; it is straightforward and characterized by low computational complexity \cite{IKOTUN2022}. \emph{K}-means aims to find \emph{k} clusters in a dataset such that the intra-cluster similarity is high and the inter-cluster similarity is low \cite{IKOTUN2022}. A cluster is expected to include a set of isolated points \cite{Jain2010}. The \emph{k}-means achieve this by computing the Euclidean distance between the data points and the cluster centers, thus forming the spherical partitioning of the data \cite{IKOTUN2022}. In other words, \emph{k}-means start with an initial cluster and then assign the clusters minimizing the squared error until the clusters are stable \cite{Jain2010}. It should be noted that high dimensionality challenges the performance of \emph{k}-means. Therefore, principal component analysis (PCA) is used to transform the data into a lower dimensional space which facilitates the detection of coherent patterns more clearly in the data \cite{Ding2004k}. PCA is a feature reduction statistical technique used to extract important information from high-dimensional data and project it into lower-dimensional space of orthogonal variables \cite{Abdi2010}. In order to identify the number of sufficient principal components, the variance ratio for each principal component is calculated. The variance ratio is equal to the ratio of the eigenvalue of a principal component to the sum of the eigenvalues of all principal components. First, the data was standardized to ensure a mean of 0 and a variance of 1. Subsequently, the standardized data was normalized to bring all variables to a comparable scale. Given that the two principal components explain more than 80\% of the data in all three datasets, PCA was then applied with two components to reduce the dimensionality of the dataset preserving the most significant patterns. This process enables clear and interpretable two-dimensional visualization of the resulting clusters.

\emph{K}-means method requires the user to specify the number of centroids before implementation; that is \emph{k} value. Selecting the optimal \emph{k} value is not straightforward and is considered one of the most challenging steps in partitional algorithms \cite{IKOTUN2022}. Often, different \emph{k} values are tested to find the best value of \emph{k}. In this use case, \emph{k} was chosen according to the elbow method, where different \emph{k} values are tested iteratively until the best value of \emph{k} is selected. In our case, both the elbow method and silhouette analysis were used from the scikit-learn library in Python 3. The elbow method plots the within-cluster sum of squares against different values of k, where the optimal k is indicated by an "elbow" in the curve. For all three datasets, the optimal \emph{k} value is three. The results are then validated by the silhouette analysis method, where a score is assigned to how dense and well-separated a cluster is from the others. Dividing the data into three clusters, results in an average silhouette score ranging between 0.61 and 0.66 for all three datasets. Therefore, it is reasonable to assume that the data could be grouped into three clusters in each of the G1, G2, and G3 datasets. Given the exploratory nature of \emph{k}-means, we aim to understand the general statistical patterns in the clusters that can influence the ML model. 

In this case study, the application of the DQ evaluation framework includes the systematic implementation of these selected methods to all three datasets. However, the application of the DQ evaluation framework to other case studies may require using a different set of tools depending on the nature of the data. For the purpose of this use case, the proposed approach is followed in a similar way across the three datasets and is summarized in the following list of steps:

\begin{enumerate}
  \item Normalize the data and transform them into two-dimensional space through the application of PCA.
  \item Pre-identify the number of clusters (\emph{k}) through the elbow method.
  \item Validate k through the silhouette analysis method.
  \item Given k, apply the \emph{k}-means clustering method to each dataset.
  \item Plot the clusters and visualize the boundaries.
  \item Predict $t_\mathrm{R}$ in each cluster using a hyper-tuned ML model. In Table \ref{tbl:hypertuning} the hypertuning parameter grids for the different ML models are shown.
  \item Observe the statistical characteristics of the data in each cluster and infer valuable insights to improve the performance of the ML system.
\end{enumerate}

\begin{table}
\caption{Hypertuning search space for models.}
\label{tbl:hypertuning}
\centering
\begin{tabular}{ll}
\hline
\textbf{Model} & \textbf{Parameter Grid} \\
\hline
GB & \parbox{8cm}{
\begin{itemize}
\item 'max\_depth': [5, 10, 15, 20, 50]
\item 'learning\_rate': [0.001, 0.01, 0.1, 0.2]
\item 'n\_estimators': [100, 500, 1000]
\item 'max\_leaf\_nodes': [2, 5, 10]
\end{itemize}
} \\
\hline
SVR & \parbox{8cm}{
\begin{itemize}
\item 'C': [1., 21.38, 62.16, ...1000.]
\item 'gamma': [0.1, 0.01, 0.001]
\item 'epsilon': [0.0001, 0.00011111, 0.00012222, 0.0002]
\item 'kernel': 'rbf'
\end{itemize}
} \\
\hline
\end{tabular}
\end{table}

\section{Results}\label{sec:results}
\subsection{G1 Dataset}

\begin{figure*}
  \centering
  \begin{subfigure}[b]{0.6\textwidth}
    \includegraphics[width=\textwidth]{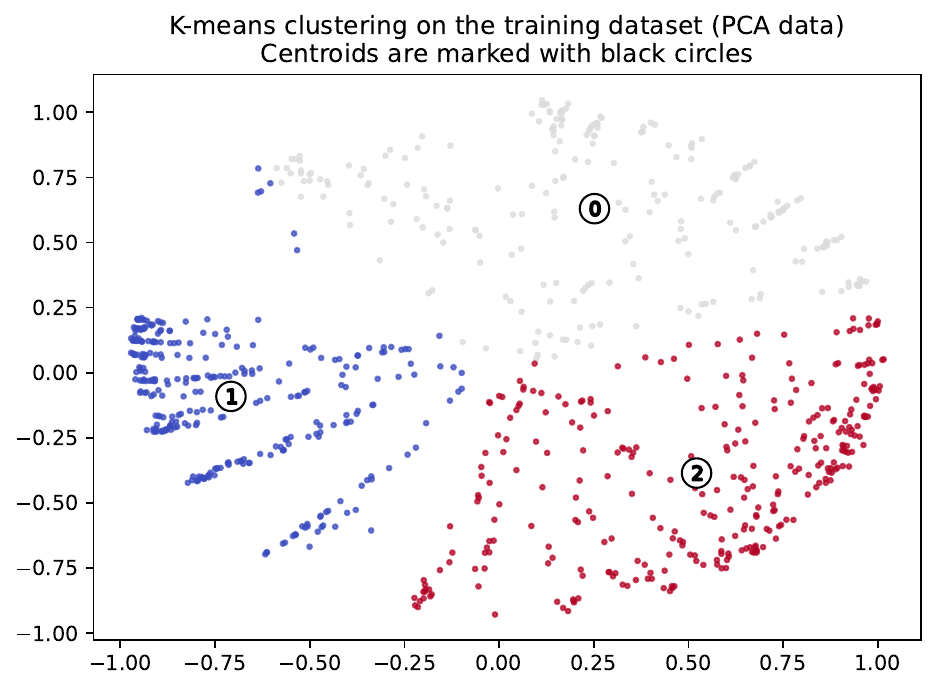}
    \caption{\emph{k}-means clusters in G1 dataset.}
  \end{subfigure}
  \par\bigskip
  \begin{subfigure}[b]{0.33\textwidth}
    \includegraphics[width=\textwidth]{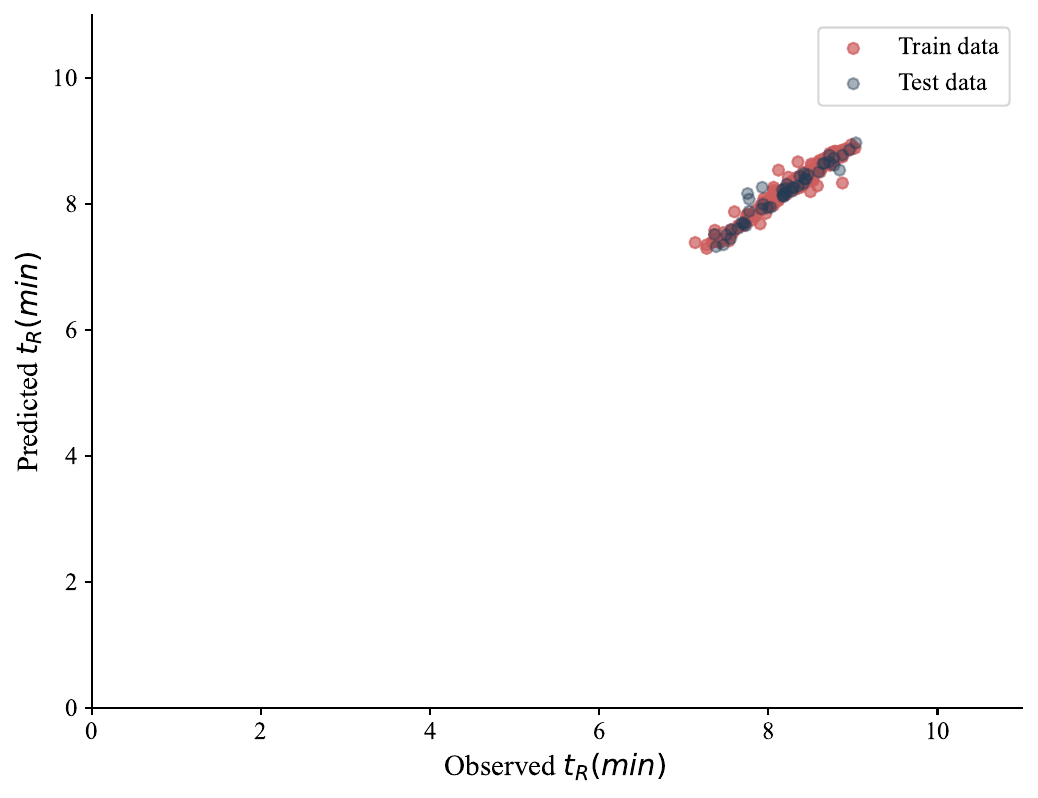}
    \caption{Cluster 0}
  \end{subfigure}
  \begin{subfigure}[b]{0.33\textwidth}
    \includegraphics[width=\textwidth]{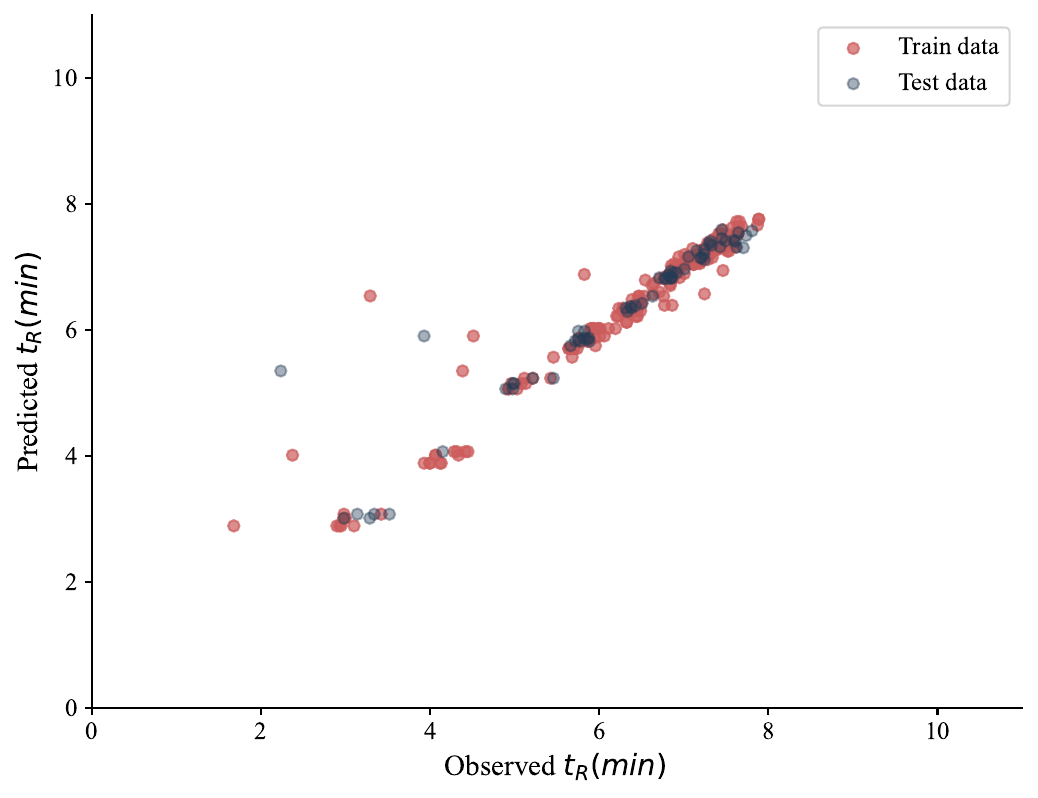}
    \caption{Cluster 1}
  \end{subfigure}
  \begin{subfigure}[b]{0.33\textwidth}
    \includegraphics[width=\textwidth]{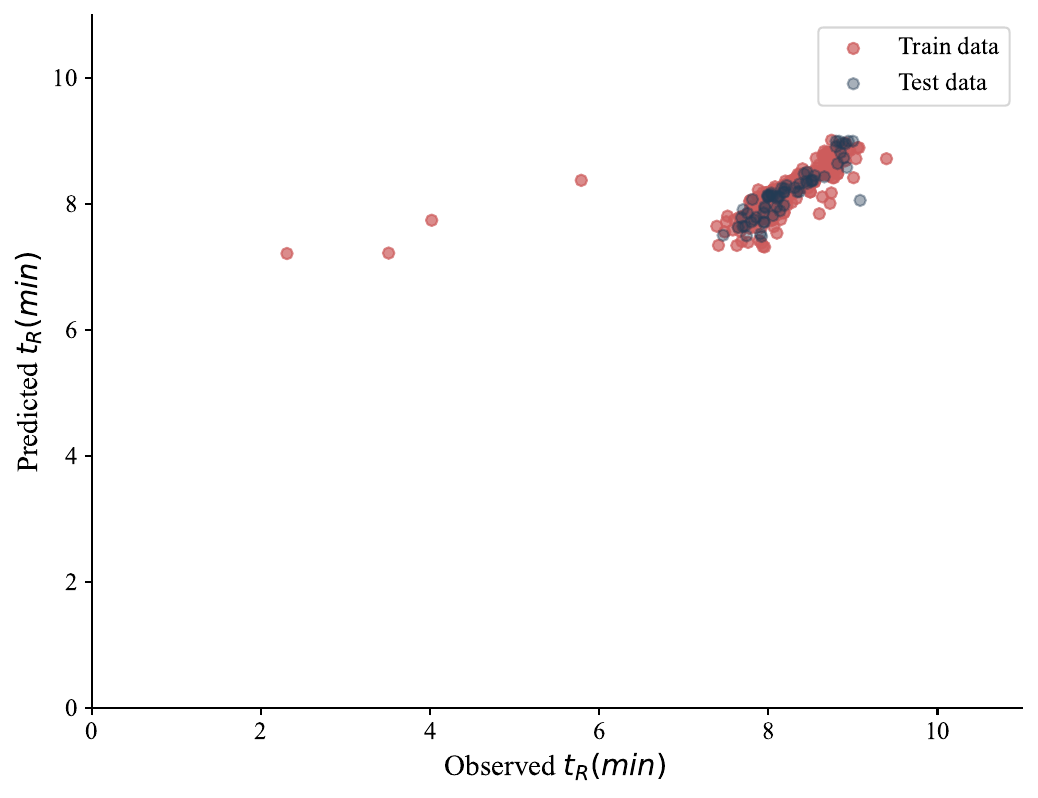}
    \caption{Cluster 2}
  \end{subfigure}

\caption{Observed versus predicted $t_\mathrm{R}$ for G1 dataset.}
  \label{fig:G1clusters}
\end{figure*}

The quality-centric data evaluation framework is tested on the G1 dataset resulting in the grouping of the data into three clusters as shown in Figure \ref{fig:G1clusters}(a). The results of applying the hypertuned Gradient Boost model to each cluster are visualized in Figures~\ref{fig:G1clusters}(b-d), where cluster 0 shows the best performance with tightly grouped predictions around the ideal line (Figure~\ref{fig:G1clusters}(b)), while cluster 2 displays more scattered predictions (Figure~\ref{fig:G1clusters}(d)).

The statistical characteristics of each cluster are presented in Tables \ref{tab:G1C0stats}, \ref{tab:G1C1stats}, and \ref{tab:G1C2stats}. After the data were clustered according to the quality measurements, the hypertuned Gradient Boost model was applied to the three clusters. The result of the model performance in the G1 test set is shown in Table~\ref{tab:G1ML}. The model performed poorly in cluster 2, unlike clusters 0 and 1, where R\textsuperscript{2} recorded 0.95 and 0.78, respectively. The Gradient boost model performed best on cluster 0.

\begin{table}
  \centering
  \caption{Evaluation of the performance of ML models in the three clusters in G1. The root mean square error (RMSE) train and R\textsuperscript{2} train in G1 are 0.368 and 0.906, respectively.}
  \label{tab:G1ML}
  \begin{tabular}{
    c 
    S[table-format=2.2] 
    S[table-format=1.2] 
  }
    \toprule
    {Cluster\#} & {RMSE test} & {R\textsuperscript{2} test} \\
    \midrule
     Cluster 0 & 0.08 & 0.95 \\
     Cluster 1 & 0.57 & 0.78 \\
     Cluster 2 &  0.71 & 0.03 \\
    \bottomrule
  \end{tabular}
\end{table}

Table \ref{tab:G1C0stats} shows the characteristics of cluster 0 in the G1 dataset. The cluster consists of 237 ASOs. The mean $t_\mathrm{R}$ difference, represented by $\Delta$t$_R$, between the two runs is 0.07 minutes. The average SNR is 603.82, significantly higher than clusters 1 and 2. The variation in the SNR values is high, with a maximum value of 8155.01. On average, the ASOs in this cluster are 15.93 long and modified with at most 11 sulfur atoms. Half of the sequences in cluster 0 are non-phosphorothioated sequences with an average of $t_\mathrm{R}$ 8.22 minutes. 

\begin{table}
\setlength{\tabcolsep}{3 pt}
\centering
\caption{Statistical characteristics of data in cluster 0 (237 ASO) in the G1 dataset. \\Not applicable values in the Injection Volume feature are represented by a dashed line ($-$).}
\label{tab:G1C0stats}
\resizebox{\textwidth}{!}{
\begin{tabular}{llllllllc}
\toprule
{} & \textbf{{$\Delta$t$_R$}} & \textbf{SNR} & \textbf{Skewness} & \textbf{Peak area} & \textbf{Length} & \textbf{Sulfur\#}& \textbf{$t_{R}$ (\si{\minute})} & \textbf{Injection} \\
&&&&&&&& \textbf{Volume (\si{\micro\liter})} \\
\midrule
\textbf{mean} &0.07& 603.82& 1.34& 19337.86& 15.93&	2.79 & 8.22&$-$ \\
\textbf{std} & 0.76& 1260.8& 0.30& 51767.07&	2.54 & 3.43& 0.43&$-$ \\
\textbf{min} &$-0.92$&	1.64&	0.52 & 9.52&	12&	0& 7.13& 0.5\\
\textbf{25\%} &$-0.01$&	32.58&	1.13&	448.81&	14&	0&7.94& $-$ \\
\textbf{median} & $-0,01$&	123.26&	1.27&	2150.91&	16&	0&8.26&$-$ \\
\textbf{75\%} &1E-04& 462.95& 1.47&	9550.88& 18& 6&8.57&$-$ \\
\textbf{max} &6.0&	8155.01&	2.54&	406506.31& 20& 11&9.03& 25\\
\bottomrule
\end{tabular}
}
\end{table}

By observing Table \ref{tab:G1C0stats} and Table \ref{tab:G1C1stats} representing the first two clusters, we notice significant differences in the values of the SNR, Length, and Sulfur\# features. Cluster 1 has 313 ASO, on average 9.64 long and likely to contain sulfur. On average, cluster 1 has a lower SNR and higher skewness than cluster 0, indicating that the peaks in cluster 1 are slightly more skewed. The mean $t_{R}$ value for cluster 0 is 8.22 minutes, while for cluster 1, it is 6.42 minutes. Sequences in this cluster have a shorter $t_\mathrm{R}$ than those in cluster 0.

\begin{table}
\setlength{\tabcolsep}{3 pt}
\centering
\caption{Statistical characteristics of data in cluster 1 (313 ASO) in the G1 dataset. \\Not applicable values in the Injection Volume feature are represented by a dashed line ($-$).}
\label{tab:G1C1stats}
\resizebox{\textwidth}{!}{
\begin{tabular}{llllllllc}
\toprule
{} & \textbf{{$\Delta$t$_R$}} & \textbf{SNR} & \textbf{Skewness} & \textbf{Peak area} & \textbf{Length} & \textbf{Sulfur\#}& \textbf{$t_{R}$ (\si{\minute})} & \textbf{Injection} \\
&&&&&&&& \textbf{Volume (\si{\micro\liter})} \\
\midrule
\textbf{mean} &0.10& 206.35& 1.52& 5383.54&	9.64& 5.79&6.42& $-$ \\
\textbf{std} & 1.09& 330.0&	0.66& 10710.67&	1.98& 3.58&1.21&$-$ \\
\textbf{min} &$-7,6172$& 2.71&	0.44&	19.33&	5&	0& 1.67&0.5\\
\textbf{25\%} &$-0.01$&	32.25& 1.22& 450.79& 8&	3&5.89 &$-$ \\
\textbf{median} &$-0.00015$& 85.78&	1.33& 1943.75&	10&	6& 6.83&$-$ \\
\textbf{75\%} &0.006& 268.10 &	1.53&	6356.90 &	11&	9&7.27&$-$ \\
\textbf{max} &5.88&	3421.18& 5.87&	133147.98& 13	&11 & 7.88 & 25\\
\bottomrule
\end{tabular}
}
\end{table}

The statistics of the third cluster in the G1 dataset are presented in Table \ref{tab:G1C2stats}. The average value of $\Delta t_R$ in cluster 2 is lower than that of the other two clusters, similarly, for the average SNR and the peak area values. The average skewness value is 1.44, larger than the average value for cluster 0 and smaller than the average value for cluster 0. The ASO sequence is partially or fully phosphorothioated and is at least 9 nucleotides long.

\begin{table}
\setlength{\tabcolsep}{3 pt}
\centering
\caption{Statistical characteristics of data in cluster 2 (315 ASO) in the G1 dataset. \\Not applicable values in the Injection Volume feature are represented by a dashed line ($-$).}
\label{tab:G1C2stats}
\resizebox{\textwidth}{!}{
\begin{tabular}{llllllllc}
\toprule
{} & \textbf{{$\Delta$t$_R$}} & \textbf{SNR} & \textbf{Skewness} & \textbf{Peak area} & \textbf{Length} & \textbf{Sulfur\#}& \textbf{$t_{R}$ (\si{\minute})} & \textbf{Injection} \\
&&&&&&&& \textbf{Volume (\si{\micro\liter})} \\
\midrule
\textbf{mean} &0.03&	164.58&	1.44&	3500.56&	15.99&	14.26 & 8.19& $-$ \\
\textbf{std} & 0.91& 352.19& 0.42&	9063.95&	2.17&	2.38 & 0.63 &$-$ \\
\textbf{min} &$-5.14$&	2.76&	0.53&	9.78&	13&	9 &2.30&0.5\\
\textbf{25\%} &$-0,018$	& 19.61& 1.18&	216.95&	14	& 13 & 7.95 & $-$ \\
\textbf{median} &1.66E-05&	45.24&	1.34&	670.10&	16&	14& 8.18 & $-$ \\
\textbf{75\%} &0.01 & 157.44 & 1.58 &	2646.50 & 18&	16& 8.54 & $-$ \\
\textbf{max} &6.40& 2484.56&	3.43&	77796.73&	20&	19&	9.39 & 25\\
\bottomrule
\end{tabular}
}
\end{table}

\subsection{G2 Dataset}

\begin{figure*}
  \centering
  \begin{subfigure}[b]{0.6\textwidth}
    \includegraphics[width=\textwidth]{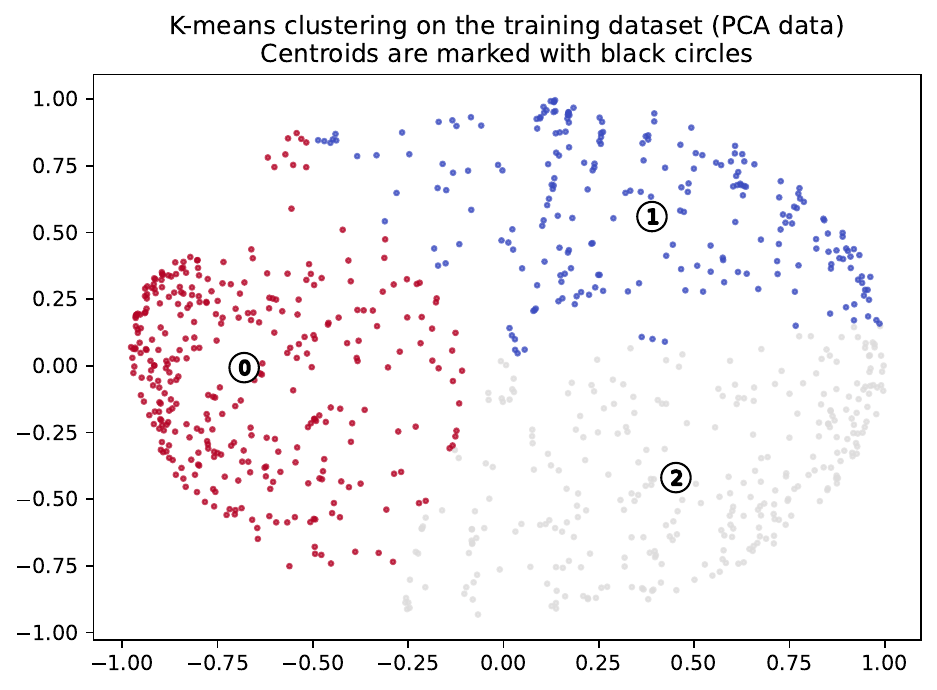}
    \caption{\emph{k}-means clustering in G2 dataset.}
  \end{subfigure}
  \par\bigskip
  \begin{subfigure}[b]{0.33\textwidth}
    \includegraphics[width=\textwidth]{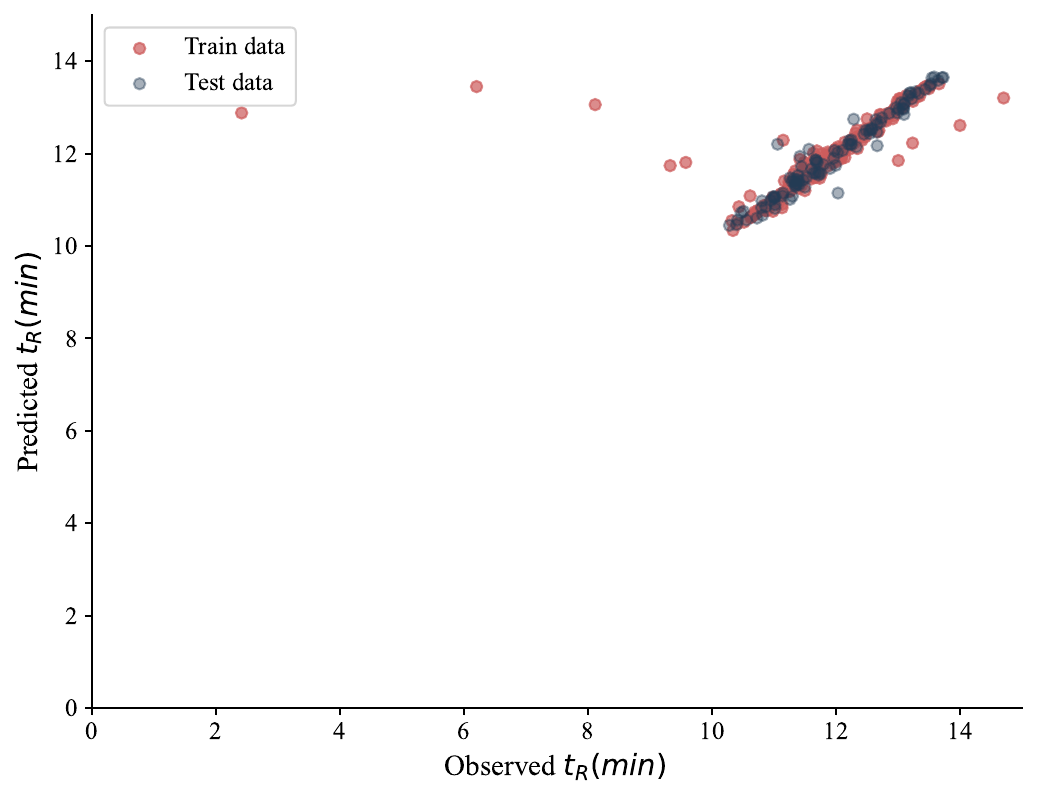}
    \caption{Cluster 0}
  \end{subfigure}
  \begin{subfigure}[b]{0.33\textwidth}
    \includegraphics[width=\textwidth]{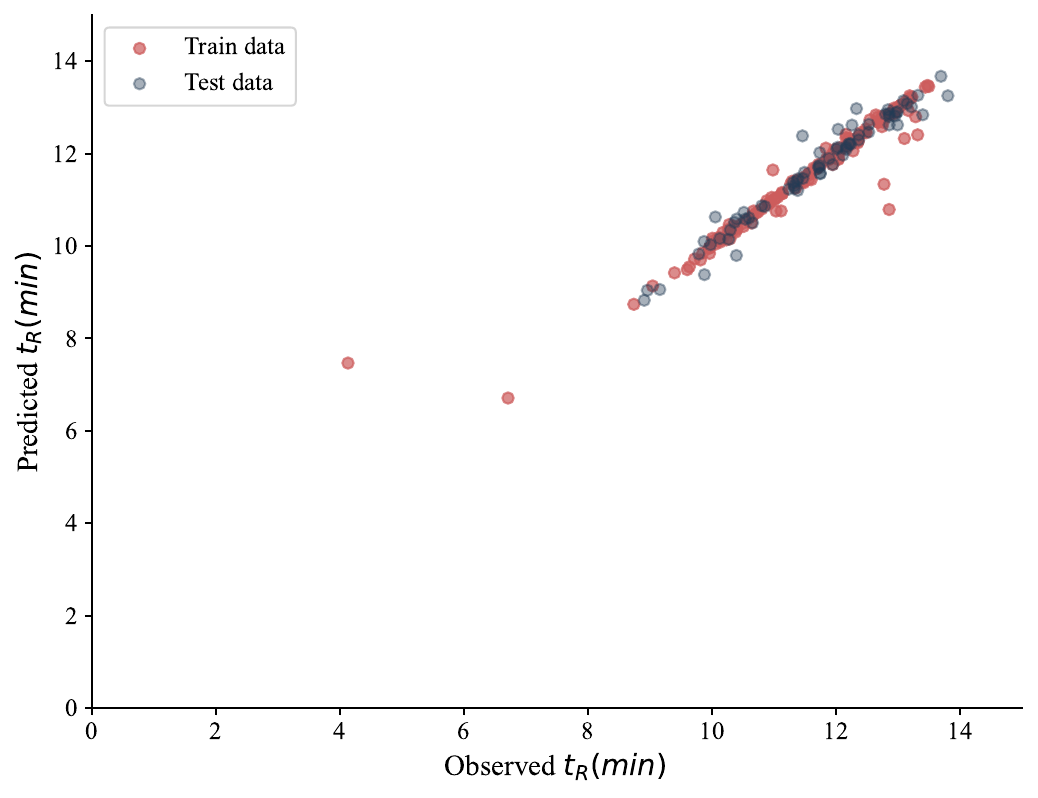}
    \caption{Cluster 1}
  \end{subfigure}
  \begin{subfigure}[b]{0.33\textwidth}
    \includegraphics[width=\textwidth]{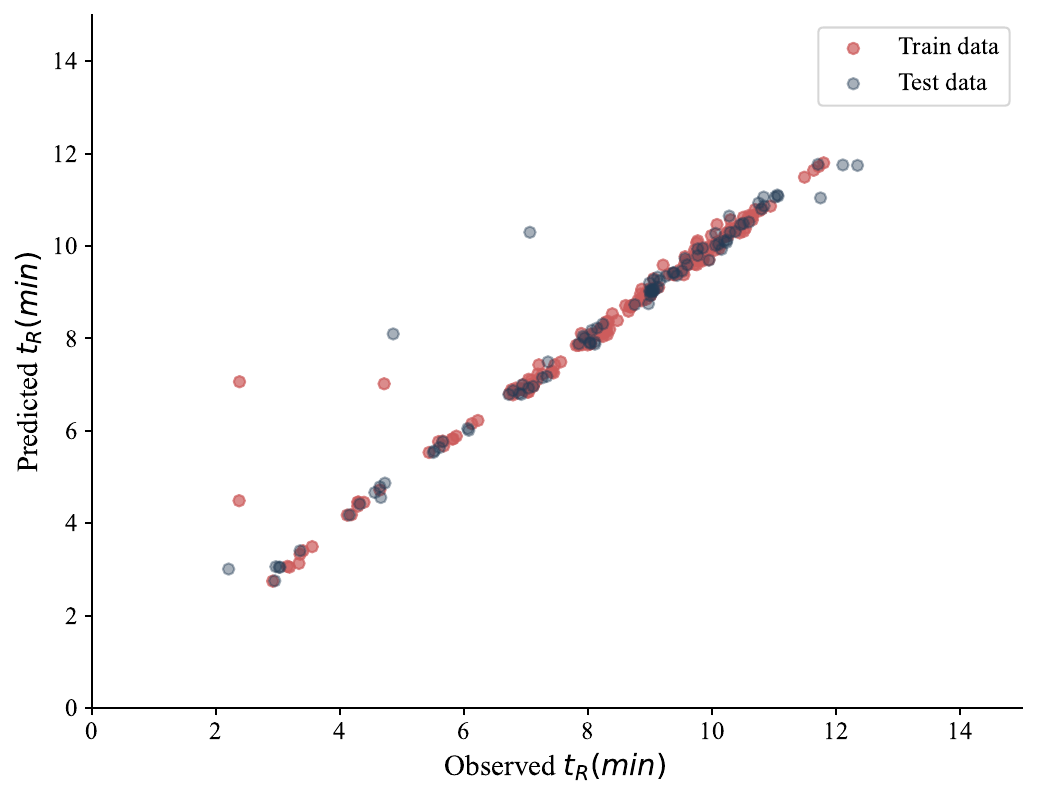}
    \caption{Cluster 2}
  \end{subfigure}
  \caption{Observed versus predicted t$_\mathrm{R}$ for G2 dataset.}
  \label{fig:G2clusters}
\end{figure*}

The result of the application of the hypertuned support vector regression (SVR) model to the three identified clusters in G2 is summarised in Table~\ref{tab:G2ML} and Figure~\ref{fig:G2clusters}. The clustering results visualized in Figure~\ref{fig:G2clusters}(a) show the distribution of data points in the reduced dimensional space, with clear separation between quality-based groups. Table~\ref{tab:G2ML} shows the RMSE and R\textsuperscript{2} values in G2 test sets for each of the three clusters. According to the results, the performance of the SVR model varies across the three clusters, as demonstrated in Figures~\ref{fig:G2clusters}(b-d). Cluster 0 has the highest RMSE and the lowest R\textsuperscript{2} in the test set, with scattered predictions shown in Figure~\ref{fig:G2clusters}(b), indicating poor performance in this cluster compared to the other clusters. Cluster 1 and cluster 2 have lower RMSE and higher R\textsuperscript{2} values for both train and test sets, as evidenced by the tighter grouping of predictions around the ideal line in Figures~\ref{fig:G2clusters}(c,d), with SVR recording the lowest RMSE in cluster 1. With respect to RMSE, the performance of the SVR shows to be better in cluster 1 compared to the other clusters.

\begin{table}
  \centering
  \caption{Evaluation of the performance of ML models in the three clusters in G2. The  RMSE train and R\textsuperscript{2} train in G2 are 0.641 and 0.917 respectively.}
  \label{tab:G2ML}
  \begin{tabular}{
    c 
    S[table-format=2.2] 
    S[table-format=1.2] 
  }
    \toprule
    {Cluster\#} & {RMSE test} & {R\textsuperscript{2} test} \\
    \midrule
     Cluster 0 & 1.122 & 0.282 \\
     Cluster 1 & 0.598 & 0.855 \\
     Cluster 2 & 0.701 & 0.889 \\
    \bottomrule
  \end{tabular}
\end{table}

The first cluster in the G2 dataset has 331 ASO compounds. In cluster 0, the $\Delta$t$_R$ is of mean equals 0.18 minutes and standard deviation of 1.78 minutes. The minimum and maximum values are $-9.75$ and 11.17 minutes respectively, which shows high variation among the two runs. The minimum and maximum values in the SNR column are 2.04 and 3004.23, respectively, where half of the data having SNR equal 44.04. As for skewness, which measures the asymmetry of the peak shape, minimum and maximum values are 0.44 and 3.33, respectively, and a mean of 1.51. The variation in the peak area data is the highest, where the standard deviation is 10736.44 and the mean equals 3974.07. Half of the peaks in this cluster have a peak area below 707.31, which could indicate the presence of outliers. The length of the sequences belonging to this cluster ranges between 13 and 20 nucleotides long, with a mean equal to 16.19 nucleotides long. As shown in Table~\ref{tab:G2C0stats}, the ASOs in cluster 0 are all phosphorothioated with at least five sulfur atoms.

\begin{table}
\setlength{\tabcolsep}{3 pt}
\centering
\caption{Statistical characteristics of data in cluster 0 (331 ASO) in the G2 dataset. \\Not applicable values in the Injection Volume feature are represented by a dashed line ($-$).}
\label{tab:G2C0stats}
\resizebox{\textwidth}{!}{
\begin{tabular}{llllllllc}
\toprule
{} & \textbf{{$\Delta$t$_R$}} & \textbf{SNR} & \textbf{Skewness} & \textbf{Peak area} & \textbf{Length} & \textbf{Sulfur\#}& \textbf{$t_{R}$ (\si{\minute})} & \textbf{Injection} \\
&&&&&&&& \textbf{Volume (\si{\micro\liter})} \\
\midrule
\textbf{mean} & 0.18 & 165.77 & 1.51 & 3974.07 & 16.19 & 13.88 & 11.95 & $-$ \\
\textbf{std} & 1.78 & 384.74 & 0.48 & 10736.44 & 2.22 & 2.82 & 1.08 & $-$ \\
\textbf{min} & $-9.75$ & 2.04 & 0.44 & 9.98 & 13 & 5 & 2.41 & 0.5\\
\textbf{25\%} & $-0.01$ & 18.78 & 1.16 & 187.72 & 14 & 12 & 11.34& $-$ \\
\textbf{median} & 0.00005 & 44.04 & 1.38 & 707.31 & 16 & 14 & 11.90 & $-$ \\
\textbf{75\%} & 0.01 & 151.10 & 1.75 & 2776.96 & 18 & 16 & 12.71 & $-$ \\
\textbf{max} & 11.17 & 3004.23 & 3.33 & 94589.24 & 20 & 19 & 14.70 & 25\\
\bottomrule
\end{tabular}
}
\end{table}

Table \ref{tab:G2C1stats} presents the statistical characteristics of the second cluster in the G2 dataset. In total, 231 data points are present in this cluster, and their lengths range between 8 and 20 nucleotides. The majority of the ASO compounds in this cluster are non-phosphorothioated. The mean $\Delta$t$_R$ is 0.26, and the standard deviation is 1.52 minutes. The negative values in $\Delta$t$_R$ show that some samples had higher t$_\mathrm{R}$ in the second run. The 25th percentile is $-0.01$ minutes, indicating that 25\% of the ASO compounds in this cluster had a higher t$_\mathrm{R}$ in the second run. The minimum SNR value is 2.36, and the maximum value is 8129.39 indicating a wide range of values. The skewness characteristic indicates that the distribution is slightly skewed to the right since the 25th percentile is greater than 1. The minimum and maximum peak area values are 13.11 and 346562.78, indicating a significant variation in the size of the peaks in this cluster. The mean t$_{R}$ is 11.52 minutes.

\begin{table}
\setlength{\tabcolsep}{3 pt}
\centering
\caption{Statistical characteristics of data in cluster 1 (231 ASO) in the G2 dataset. \\Not applicable values in the Injection Volume feature are represented by a dashed line ($-$).}
\label{tab:G2C1stats}
\resizebox{\textwidth}{!}{
\begin{tabular}{llllllllc}
\toprule
{} & \textbf{{$\Delta$t$_R$}} & \textbf{SNR} & \textbf{Skewness} & \textbf{Peak area} & \textbf{Length} & \textbf{Sulfur\#} & \textbf{$t_{R}$ (\si{\minute})} & \textbf{Injection} \\
&&&&&&&& \textbf{Volume (\si{\micro\liter})} \\
\midrule
\textbf{mean} & 0.26 & 688.63 & 1.25 & 25238.18 & 14.93 & 2.33 & 11.52 & $-$ \\
\textbf{std} & 1.52 & 1344.77 & 0.28 & 59248.99 & 2.85 & 3.33 & 1.28 & $-$ \\
\textbf{min} & $-4.29$ & 2.36 & 0.21 & 13.11 & 8 & 0 & 4.12 & 0.5\\
\textbf{25\%} & $-0.01$ & 48.31 & 1.08 & 892.95 & 12 & 0 & 10.61 & $-$ \\
\textbf{median} & 0.000016 & 164.39 & 1.21 & 4006.56 & 15 & 0 & 11.72  & $-$ \\
\textbf{75\%} & 0.01 & 602.41 & 1.38 & 16410.98 & 17 & 5 & 12.43 & $-$ \\
\textbf{max} & 10.74 & 8129.39 & 2.19 & 346562.78 & 20 & 12 & 13.80 & 25 \\
\bottomrule
\end{tabular}
}
\end{table}

Table~\ref{tab:G2C2stats} presents the statistical characteristics of the third and last cluster in the G2 dataset. The mean t$_{R}$ is 8.40 minutes, slightly less than the average t$_\mathrm{R}$ in the other two clusters. The average SNR is 139.15, indicating a moderate level of the SNR ratio. The average skewness value is 1.64, showing a relatively asymmetric distribution of the peaks. The average area is 3963.97, with a standard deviation of 5735.66, indicating a large amount of variability in the peak area data. The ASO in cluster 2 is a mixture of phosphorothioated and non-phosphorothioated sequences with at most 12 sulfur atoms.

\begin{table}
\setlength{\tabcolsep}{3 pt}
\caption{Statistical characteristics of data in cluster 2 (301 ASO) in the G2 dataset.\\Not applicable values in the Injection Volume feature are represented by a dash ($-$).}
\label{tab:G2C2stats}
\resizebox{\textwidth}{!}{
\begin{tabular}{llllllllc}
\toprule
{ } & {$\Delta$t$_R$} & \textbf{SNR} & \textbf{Skewness} & \textbf{Peak area} & \textbf{Length} & \textbf{Sulfur\#} & \textbf{$t_{R}$ (\si{\minute})} & \textbf{Injection} \\
&&&&&&&& \textbf{Volume (\si{\micro\liter})} \\
\midrule
\textbf{mean} & $-0.01$ & 139.15 & 1.64 & 3963.97 & 9.75 & 6.10 & 8.40 & $-$ \\
\textbf{std} &2.06 & 173.98 & 0.56 & 5735.66 & 2.20 & 3.50 & 2.11 & $-$ \\
\textbf{min} & $-12.3$ & 4.22 & 0.62 & 8.31 & 5 & 0 & 2.20 & 0.5 \\
\textbf{25\%} & $-0.03$ & 27.06 & 1.30 & 397.59 & 8 & 3 & 7.12 & $-$ \\
\textbf{median} & $-0.00015$ & 66.44 & 1.46 & 1337.53 & 10 & 7 & 8.97 & $-$ \\
\textbf{75\%} & 0.01 & 196.76 & 1.80 & 5517.71 & 11 & 9 & 9.90 & $-$ \\
\textbf{max} & 8.84 & 1318.21 & 4.35 & 38255.22 & 16 & 12 &12.34 & 25\\
\bottomrule
\end{tabular}
}
\end{table}

\subsection{G3 Dataset}

\begin{figure*}
  \centering
  \begin{subfigure}[b]{0.6\textwidth}
    \includegraphics[width=\textwidth]{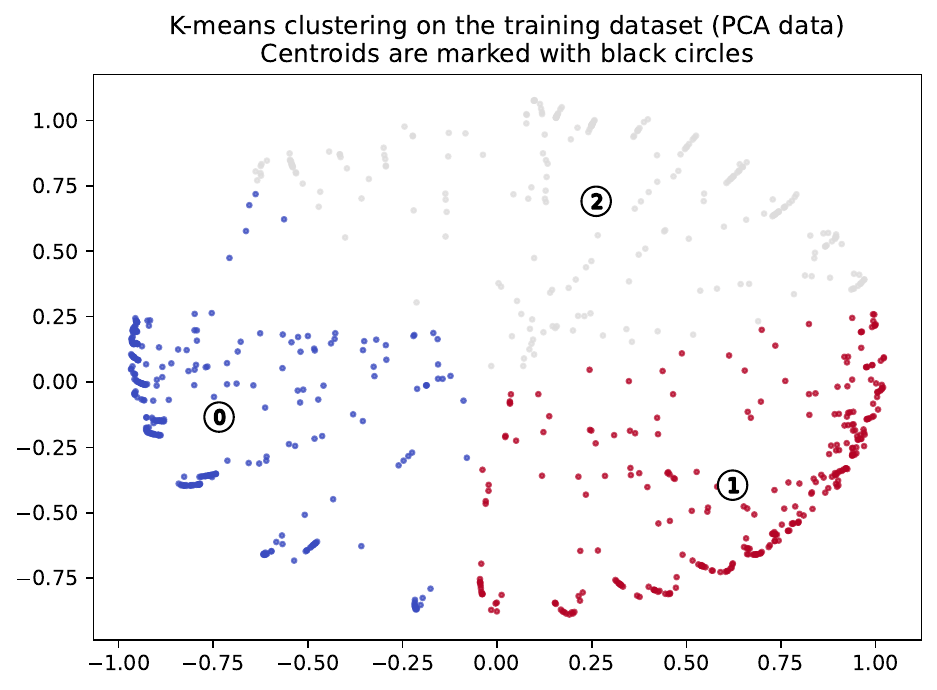}
    \caption{\emph{k}-means clustering in G3 dataset.}
  \end{subfigure}
  \par\bigskip
  \begin{subfigure}[b]{0.33\textwidth}
    \includegraphics[width=\textwidth]{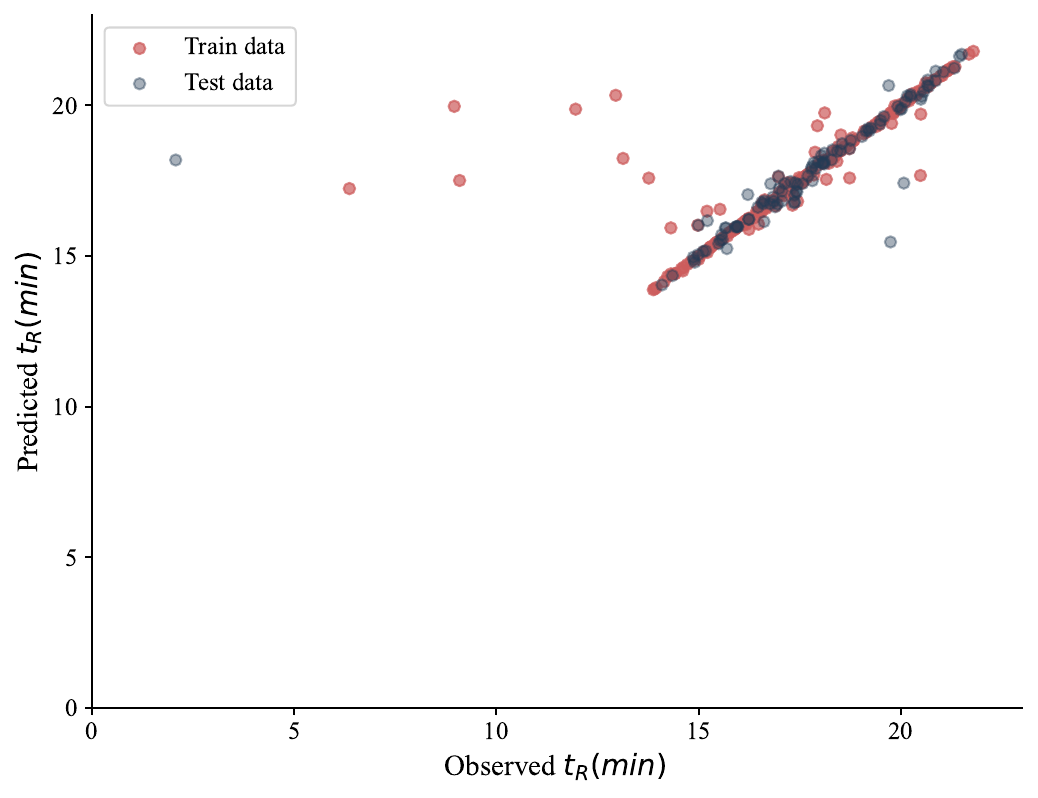}
    \caption{Cluster 0}
  \end{subfigure}
  \begin{subfigure}[b]{0.33\textwidth}
    \includegraphics[width=\textwidth]{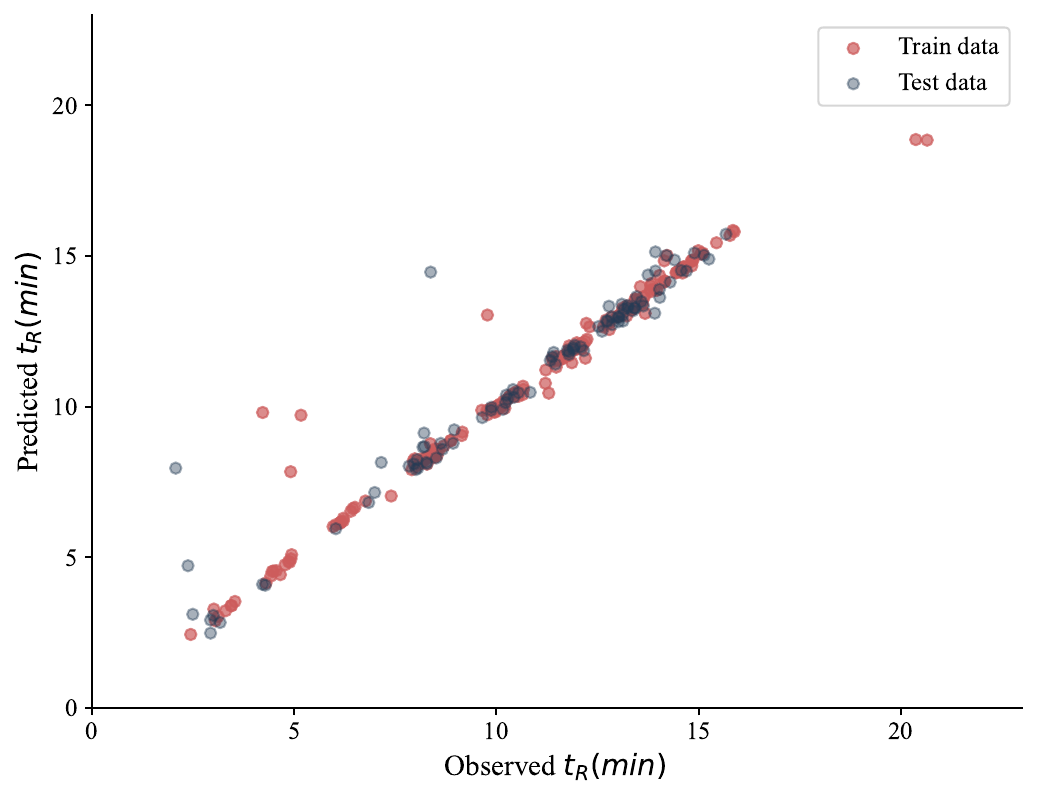}
    \caption{Cluster 1}
  \end{subfigure}
  \begin{subfigure}[b]{0.33\textwidth}
    \includegraphics[width=\textwidth]{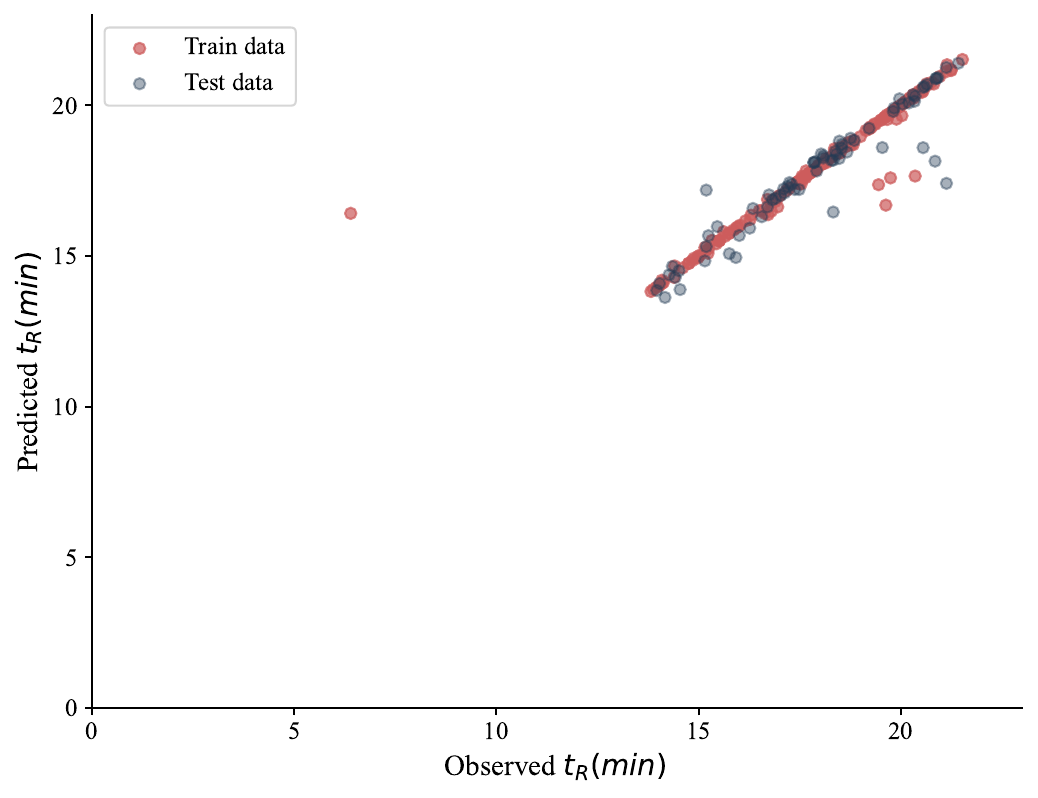}
    \caption{Cluster 2}
  \end{subfigure}
  \caption{Observed versus predicted t$_\mathrm{R}$ for G3 dataset.}
  \label{fig:G3clusters}
\end{figure*}

G3 is the most sensitive dataset among all datasets, with the highest noise. Similar to the G1 and G2 datasets, the quality-centric data evaluation framework was applied to the G3 dataset, resulting in the clustering of the data into three groups as shown in Figure~\ref{fig:G3clusters}(a), where the PCA-transformed data points show distinct quality-based groupings. The statistical characteristics of each cluster are presented in Tables~\ref{tab:G3C0stats}, \ref{tab:G3C1stats}, and \ref{tab:G3C2stats}. In the next step, the hypertuned SVR model was applied to the three identified clusters, with the prediction results visualized in Figures~\ref{fig:G3clusters}(b-d). The results of the evaluation of the SVR model on the G3 test set are shown in Table~\ref{tab:G3ML}. The SVR model performed poorly on cluster 0, as evidenced by the scattered predictions in Figure~\ref{fig:G3clusters}(b), unlike clusters 1 and 2, where the rmse recorded 0.92 and 0.49, respectively, showing tighter prediction patterns in Figures~\ref{fig:G3clusters}(c,d). The SVR model performed best in cluster 2, which is reflected in the close alignment of predictions with the ideal line in Figure~\ref{fig:G3clusters}(d).

\begin{table}
  \centering
  \caption{Evaluation of the performance of ML models in the three clusters in G3. The RMSE train and R2 train in G3 are 1.0 and 0.92 respectively.}
  \label{tab:G3ML}
  \begin{tabular}{
    c 
    S[table-format=2.2] 
    S[table-format=1.2] 
  }
    \toprule
    {Cluster\#} & {RMSE test} & {R\textsuperscript{2} test} \\
    \midrule
     Cluster 0 & 2.180 & 0.344 \\
     Cluster 1 & 0.929 & 0.924 \\
     Cluster 2 & 0.497 & 0.942 \\
    \bottomrule
  \end{tabular}
\end{table}

The variation in the $\Delta$t$_R$ feature in the first cluster in G3 is high, as shown in Table~\ref{tab:G3C0stats}. The minimum and maximum values for $\Delta$t$_R$ are -18.67 and 16.17, respectively. The average SNR value in cluster 0 is 145.72. The minimum and maximum values for SNR are 1.54 and 3460.76, respectively, suggesting the detection of weak and stronger signals. The degree of skewness varies in this cluster, where the 25th percentile records 0.21. This suggests that the data are skewed in general. On average, ASOs have a length of 15.71 and are at least 12 nucleotides long. All sequences in this cluster are phosphorothioated with a minimum of 8 sulfur atoms.

\begin{table}
\setlength{\tabcolsep}{3 pt}
\centering
\caption{Statistical characteristics of data in cluster 0 (333 ASO) in the G3 dataset. \\Not applicable values in the Injection Volume feature are represented by a dashed line ($-$).}
\label{tab:G3C0stats}
\resizebox{\textwidth}{!}{
\begin{tabular}{llllllllc}
\toprule
{} & \textbf{{$\Delta$t$_R$}} & \textbf{SNR} & \textbf{Skewness} & \textbf{Peak area} & \textbf{Length} & \textbf{Sulfur\#}&\textbf{$t_{R}$ (\si{\minute})} & \textbf{Injection} \\
&&&&&&&& \textbf{Volume (\si{\micro\liter})} \\
\midrule
\textbf{mean} & 0.54 & 145.72& 1.67&	4287.70& 15.71&	14.08& 17.52& $-$ \\
\textbf{std} & 3.28& 364.00&0.66& 11593.51& 2.28&	2.42 &2.27 &$-$ \\
\textbf{min} &$-18.67$&	1.54& 0.21&	7.59&	12&	8& 2.06&0.5\\
\textbf{25\%} &$-0.03$&	16.60& 1.21& 129.38&	14&	12& 16.38& $-$ \\
\textbf{median} &$-8.33E-05$&	44.11& 1.54& 585.49& 15& 14& 17.48& $-$ \\
\textbf{75\%} &0.03& 151.32& 2.02& 3813.88&	18&	16& 19.04 & $-$ \\
\textbf{max} &16.17&3460.76&4.85& 98093.98& 20& 19 & 21.79& 25\\
\bottomrule
\end{tabular}
}
\end{table}

The variation in $\Delta$t$R$ is the highest cluster 1 with a standard deviation of 3.38 minutes. The average SNR is 147.97, with a standard deviation of 216.63, indicating that the ASOs in this cluster have slightly higher SNR than those in cluster 0. The standard deviation is 7240.84 indicating a high variation in the area recordings. The average length of the ASO in this cluster is 9.48 consisting of at most 13 nucleotides, significantly less than cluster 0. The ASO compound in this cluster is a mixture of non-phosphorothioates and partially phosphorothioated sequences, with at most 13 sulfur atoms. Compared to cluster 0, ASOs in cluster 1 have a smaller variation range ($\Delta$t$R$), slightly higher SNR, shorter length, and are much less phosphorothioated.

\begin{table}
\setlength{\tabcolsep}{3 pt}
\centering
\caption{Statistical characteristics of data in cluster 1 (294 ASO) in the G3 dataset. \\Not applicable values in the Injection Volume feature are represented by a dashed line ($-$).}
\label{tab:G3C1stats}
\resizebox{\textwidth}{!}{
\begin{tabular}{llllllllc}
\toprule
{} & \textbf{{$\Delta$t$_R$}} & \textbf{SNR} & \textbf{Skewness} & \textbf{Peak area} & \textbf{Length} & \textbf{Sulfur\#}& \textbf{$t_{R}$ (\si{\minute})} & \textbf{Injection} \\
&&&&&&&& \textbf{Volume (\si{\micro\liter})} \\
\midrule
\textbf{mean} & 0.33&147.97& 1.69 &4374.07&	9.48&	5.51& 10.72& $-$ \\
\textbf{std} & 3.38&	216.63&	0.70& 7240.84& 1.91&	3.43& 3.40&$-$ \\
\textbf{min} & $-16.62$&	2.49&	0.47&	8.31E-12&	5&	0 & 2.06& 0.5\\
\textbf{25\%} & $-0,03$&	25.04&	1.24&	302.82&	8&	3& 8.46&$-$ \\
\textbf{median} &$-9.16E-05$&	63.63&	1.47&	1364.47&	10&	6&11.61& $-$ \\
\textbf{75\%} & 0.03&	190.08&	1.86&	5498.89&	11&	9& 13.34&$-$ \\
\textbf{max} & 13.67& 1762.77&	4.46&	53298.57&	13&	11& 20.65& 25\\
\bottomrule
\end{tabular}
}
\end{table}

Comparing the third cluster with the two other identified clusters in the G3 dataset, it is noted that cluster 2 has the highest average SNR of 630.9 and a standard deviation of 2.67. Cluster 2, on average, is the least skewed. In this cluster, the peak area is the widest among all clusters. Overall, the ASO in cluster 2 is the longest but with the least phosphorothioated, where half of the ASOs data are not modified with sulfur.

\begin{table}
\setlength{\tabcolsep}{3 pt}
\centering
\caption{Statistical characteristics of data in cluster 2 (231 ASO) in the G3 dataset. \\Not applicable values in the Injection Volume feature are represented by a dashed line ($-$).}
\label{tab:G3C2stats}
\resizebox{\textwidth}{!}{
\begin{tabular}{llllllllc}
\toprule
{} & \textbf{{$\Delta$t$_R$}} & \textbf{SNR} & \textbf{Skewness} & \textbf{Peak area} & \textbf{Length} & \textbf{Sulfur\#}& \textbf{$t_{R}$ (\si{\minute})} & \textbf{Injection} \\
&&&&&&&& \textbf{Volume (\si{\micro\liter})} \\
\midrule
\textbf{mean}& 0.39& 630.9& 1.48& 25659.18&	15.74&	2.87 & 17.82& $-$ \\
\textbf{std} & 2.67& 1415.04& 0.64 & 63372.43& 2.57& 3.54& 2.16& $-$ \\
\textbf{min} & $-10.09$&	2.98&	0.03&	8.23& 12& 0& 6.39&0.5 \\
\textbf{25\%} & $-0.03$&	36.05&	1.11&	596.1&	14&	0& 16.42& $-$ \\
\textbf{median} &$-1E-04$& 129.27&	1.32&3015.27&	15&	0& 18.03& $-$ \\
\textbf{75\%} & 0.03& 444.68&	1.63&	12702.82&	18&	7 & 19.63& $-$ \\
\textbf{max} & 16.06& 12338.19&	6.15&	380947.23&	20&	11 &21.52& 25\\
\bottomrule
\end{tabular}
}
\end{table}

\section{Discussion}\label{sec:discussion}
The analysis of the obtained results is organized to answer the 1-3 RQs presented in Section \ref{sec:implementation}.

\subsection{RQ1: How can unsupervised learning be effective in classifying data records into different quality levels?}
The successful performance of unsupervised learning depends on the careful selection of the quality measurements that are chosen in the second stage of the DQ evaluation framework. The choice of interesting quality measurements is application-specific and depends on the successful collaboration between data scientists and domain experts. Once the clusters are generated, the dominating characteristics of the data points in each cluster can be analyzed in depth, where high- and low-quality data can be deduced. In this case study, the selection of quality measurements proved to play an important role in the clustering of ASO compounds into high- and low-quality groups.  Given that the definition of the quality measurements depends on the application being studied, it is recommended that domain experts and ML practitioners work together to identify a representative set of measurements.  

\subsection{RQ2: How do we transform the results of the application of an unsupervised DQ evaluation framework into explainable insights that improve the performance of the ML software system?}
The application of unsupervised methods alone is often recognized under the exploratory data analysis tools, which can score low on the explainability spectrum. However, when combined with predictive ML unsupervised learning can offer valuable insights. In this case study, unsupervised learning is used to group the data into quality-sensitive clusters. Thus, generating insights on the quality characteristics of the input data in each cluster. Explainability comes one step later, more specifically after predictive ML is applied. One of the approaches to achieve explainable results is by analyzing the statistical characteristics of the data present in each cluster, similar to what is done in this paper.

For instance, in the G1 dataset, the ML model performed best in cluster 0. The ASO compounds in this group are characterized by high SNR, low skewness, and relatively low standard deviation in $\Delta$ t$_R$. The ASOs in this cluster are long, consisting of a minimum of 12 nucleotides. Phosphorothioation in this dataset has negatively affected the performance of the gradient boost model, so the cluster with the majority of sulfur-modified sequences was the hardest to learn. We also note that most of the sequences in this cluster have lost a sulfur atom. Therefore, the modification by sulfur, either adding or losing a sulfur atom shows to degrade the performance of the ML model. At a higher gradient, the ML model performed best in G2 in the cluster where the ASOs are relatively long and are mostly not phosphorothioated. The distinguishing characteristic in this cluster refers to the high average SNR, which therefore contributes to a higher performance of the ML model. At the highest gradient, G3, the SVR model performed best in the cluster consisting of the least modified compounds. Similar to the G1 dataset, phosphorothioation negatively impacted the performance of the model. In summary, the results of the experiments show that high-quality ASO data are characterized by lack of or low modification, relatively long, express low values of skewness, and high values of SNR during a chromatography experiment. Through multiple discussions, the domain expert confirmed that the identified characteristics align with their understanding of the domain. This iterative validation process ensured the findings were consistent with domain knowledge, which helps ensure the framework's outputs are meaningful and can drive real-world improvements in data collection and conducting chromatography experiments.

To effectively utilize the insights learned from the analysis stage, the insights must be transformed into actionable feedback to the data source controllers. This happens by integrating a feedback mechanism between data scientists and business teams. A continuous two-way feedback mechanism improves the quality of the input data and consequently the predictive capabilities of the deployed ML model.

\subsection{RQ3: How can we validate the results of a DQ evaluation framework built on unsupervised clustering?}
The quality-centric data evaluation framework integrates an essential stage that requires applying predictive ML to the generated quality-sensitive clusters. This step is key to validating and transforming the results of unsupervised learning into quantitative metrics often selected depending on the application. The variant performance of the ML model in each of the clusters represented by quantitative statistical metrics, such as the rmse and R\textsuperscript{2} in this case study, offers a quantitative representation of the quality of the data records belonging to each cluster. For example, in the G1 dataset, the ML model performed best in cluster 0, where the performance was significantly poor in cluster 2, as shown in Table \ref{tab:G1ML}. The difference in the performance of the predictive ML model among clusters shows that high-quality data patterns have been learned from the different groups of input data. Another key step is validating the framework output with domain experts to ensure consistency of conclusions with domain knowledge, or reason about borderline cases.

\section{Threats to validity} 
The generalization of conceptual frameworks applied to specific real-world applications is often associated with threats to external validity. Such frameworks can be designed for use cases in specific domains that do not apply to others. However, the data evaluation framework presented in Figure \ref{fig:framework} could be customized at every stage of the pipeline to fit the application at hand, including the type of unsupervised and supervised learning. The framework does not offer tailored solutions to the case study but is designed to be flexible and reproducible. Furthermore, quality measurements are selected by domain experts and ML practitioners working on a specific application. The customization of the tools and methods at every stage in the framework supports the generalization intent. This applies to ML-related tasks such as the choice of models and non-ML decisions such as the choice of quality measurements. In addition, sending feedback to data source controllers for improving data acquisition and production applies to various business domains. Based on the above, the framework is a general-purpose framework that can be customized and used in different applications. An important aspect of the generalization framework is the identification of quality measurements that are application-specific and collectively determined by domain experts and ML practitioners.

On the other hand, \emph{k}-means have been reported to be sensitive to outliers in the literature. Therefore, the performance of the \emph{k}-means algorithm in the chromatography data could be influenced to some extent by the outliers in the data. The outliers in the three datasets were intentionally not processed. The purpose of not handling outliers is to study their influence on the generated clusters. Given our analysis, we believe that this framework, using the same or different ML methods, could be used in other applications.

\section{Conclusion}\label{sec:conclusion}
In this paper, we introduce the quality-centric data evaluation framework, which aims to group the data into high- and low-quality clusters. It is required to define the suitable application DQ measurements before implementing the unsupervised learning method. The framework integrates unsupervised learning to generate quality-sensitive clusters and predictive ML to validate and analyze the results. Predictive ML is applied to each resulting cluster to predict the target variable. Based on the accuracy and the performance of the ML, characteristics of high- and low-quality data are learned. High-quality ASO data are observed to have a lack of or low modification, relatively long sequence, low levels of skewness, and high SNR values. Then the deduced insights are fed back to data source controllers to quality control the operations. The framework is simple and useful in improving the outcome of the ML software system. The framework is deployed and evaluated in an analytical chemistry case study, where it proved its efficiency. In each of the three datasets, the performance of the ML model showed distinguished results among the different generated clusters. In some clusters, ML performed poorly, while in others, the performance was significantly better, which enabled learning of high-quality characteristics of an ASO compound.

The quality-centric data evaluation framework integrates general qualitative and ML methods, which facilitate the application in other domains. The user-defined measurements are application-specific, and the user can select the unsupervised or supervised ML method that best serves the application data. Therefore, this framework is presented as a general approach to selecting high-quality data that improves the performance of an ML system.

Building on these findings, several promising directions for future research emerge. Scaling the proposed framework to larger datasets across diverse domains could help validate the generalizability of the methods. Future studies might also explore leveraging emerging technologies, such as advanced deep learning approaches, to further automate DQ evaluation. However, such advancements must carefully weigh the potential benefits against practical challenges, such as the effort required to collect large datasets from chromatography experiments. Additionally, expanding the quality-centric data evaluation framework to include metrics for quantifying time and cost savings could add another valuable dimension. These enhancements would make the framework even more impactful for developing high-performing ML systems while reducing the experimental burden on domain experts across various applications.

\section{Data and Code Availability}
The data used to support the findings of this study will be made available from the corresponding author upon reasonable request.

\section{Acknowledgments}
This work was supported by the Swedish Knowledge Foundation via the KKS SYNERGY project “Improved Methods for Process and Quality Controls using Digital Tools—IMPAQCDT” (grant number 20210021). The authors of this paper thank Jakob Häggström and Patrik Forssén. All data were collected at Karlstad University.


\end{document}